\documentclass[10pt,twocolumn,letterpaper]{article}
\usepackage{cvpr}
\usepackage{times}
\usepackage{epsfig}
\usepackage{graphicx}
\usepackage{amsmath}
\usepackage{amssymb}
\usepackage{bm}
\usepackage{array,multirow}
\usepackage[font=small]{caption}
\usepackage[nocompress]{cite}

\usepackage[pagebackref=true,breaklinks=true,letterpaper=true,colorlinks,bookmarks=false]{hyperref}

\cvprfinalcopy %
\def\cvprPaperID{2375} %

\ifcvprfinal\fi

\begin{document}
\setlength{\abovedisplayskip}{3pt}
\setlength{\belowdisplayskip}{3pt}
	
\title{\ D3VO: Deep Depth, Deep Pose and Deep Uncertainty\\for 
Monocular Visual Odometry}
\author{
Nan Yang$^\text{1,2}$ \quad Lukas von Stumberg$^\text{1,2}$ \quad Rui 
Wang$^\text{1,2}$ \quad Daniel Cremers$^\text{1,2}$\\
$^\text{1}$ Technical University of Munich \quad $^\text{2}$ Artisense
}

\maketitle

\begin{abstract}
We propose D3VO as a 
novel framework for monocular visual odometry that 
exploits deep networks on three levels -- deep depth, pose and uncertainty estimation. 
We first propose a novel 
self-supervised monocular depth estimation network trained on stereo videos 
without any external supervision. In particular, it aligns the training image 
pairs into similar lighting condition with predictive brightness transformation 
parameters. Besides, we model the photometric uncertainties of pixels on 
the input images, which improves the depth estimation accuracy and 
provides a learned weighting function for the photometric residuals in direct 
(feature-less) visual odometry. Evaluation results show that the proposed 
network outperforms state-of-the-art self-supervised depth estimation networks. 
D3VO tightly incorporates the 
predicted depth, pose and uncertainty into a  direct visual odometry method to boost both 
the front-end tracking as well as the back-end non-linear optimization. We 
evaluate D3VO in terms of monocular visual odometry on both the KITTI 
odometry benchmark and the EuRoC MAV dataset.
The results show that D3VO outperforms state-of-the-art traditional 
monocular VO methods by a large margin. It also achieves comparable results to 
state-of-the-art stereo/LiDAR odometry on KITTI and to the 
state-of-the-art visual-inertial odometry on EuRoC MAV, while using 
only a single camera. 

\end{abstract}

\section{Introduction}
\begin{figure}
	\centering
	\includegraphics[width=\linewidth]{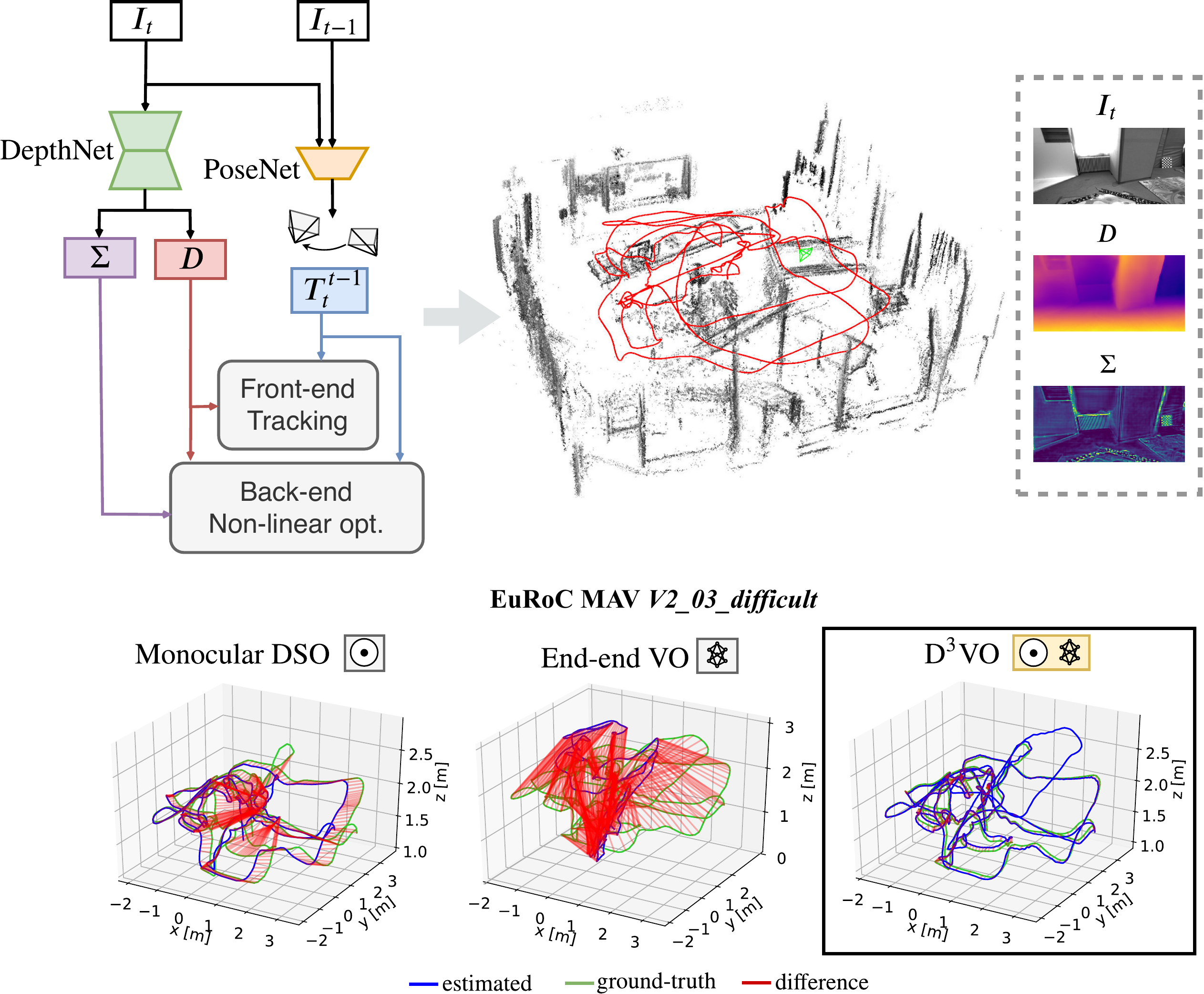}
	\caption{We propose D3VO -- a novel monocular visual 
		odometry (VO) framework which exploits deep 
		neural networks on three levels: \textbf{D}eep depth 
		($D$), \textbf{D}eep pose ($T^{t-1}_{t}$) and \textbf{D}eep 
		uncertainty ($\Sigma$) 
		estimation. 
		D3VO integrates the three estimations tightly into both the 
		front-end tracking and the back-end non-linear optimization of 
		a sparse direct odometry framework~\cite{engel2017direct}.}\vspace{-1.5em}
	\label{fig:teaser}
\end{figure}
Deep learning has swept most areas of computer vision -- not only 
high-level tasks 
like object classification, detection and 
segmentation~\cite{ren2015faster,he2017mask,kirillov2019panoptic},
but also low-level ones such as 
optical flow estimation~\cite{dosovitskiy2015flownet,pwcnet} and interest point
detection and description~\cite{yi2016lift,detone2018superpoint,dusmanu2019d2}. Yet, 
in the field 
of Simultaneously Localization And Mapping (SLAM) or Visual Odometry 
(VO) which estimates 
the relative camera poses from image sequences, traditional 
geometric-based 
approaches~\cite{mur2017orb,engel2014lsd,engel2017direct} still 
dominate the 
field. While monocular methods~\cite{mur2015orb,engel2017direct} have 
the 
advantage of low hardware 
cost and less calibration effort, they cannot achieve 
competitive performance compared to 
stereo~\cite{mur2017orb,wang2017stereoDSO} or visual-inertial odometry 
(VIO)~\cite{von2018direct,leutenegger2015keyframe,qin2018vins,mur2017visual},
due to the scale drift~\cite{strasdat2010scale,yang2018challenges} and low 
robustness. 
Recently, there have been many efforts to address this by leveraging 
deep neural 
networks~\cite{tateno2017cnn,loo2019cnn,zhan2019visual,yin2017scale}.
 It has been shown that with deep monocular depth estimation 
networks~\cite{godard2016unsupervised,Godard_2019_ICCV,laina2016deeper, yang2018deep},
the performance of monocular VO is boosted, since deep networks are 
able to estimate depth maps with consistent metric scale by learning 
a-priori knowledge from a large amount of 
data~\cite{kuznietsov2017semi}. 

In this way, however, deep neural networks are only used to a limited 
degree. Recent advances of self- and unsupervised monocular depth 
estimation networks~\cite{zhou2017unsupervised,Godard_2019_ICCV} show 
that the poses of the adjacent monocular frames can be predicted 
together with the depth. Since the pose estimation from deep neural 
networks shows high robustness, one question arises: \textit{Can the 
deep-predicted poses be employed to boost traditional VO}? 
On the other hand, since SLAM/VO is essentially a state estimation problem where 
uncertainty plays an important 
role~\cite{thrun2005probabilistic,engel2013iccv,whyfilter} and meanwhile many learning based
methods have started estimating uncertainties, the next question is, \textit{how 
can we incorporate such uncertainty-predictions into optimization-based VO}?  

In this paper, we propose D3VO as a framework for monocular direct 
(feature-less) visual VO that exploits self-supervised 
monocular depth estimation network on three levels: \textit{deep 
depth}, \textit{pose} and \textit{uncertainty} estimation, as shown in 
Fig.~\ref{fig:teaser}. To this end, 
we first propose a purely self-supervised network trained with stereo 
videos. The proposed self-supervised network predicts the depth from a 
single image with DepthNet and the pose between two adjacent frames 
with PoseNet. The two networks are bridged by minimizing the 
photometric error originated from both \textit{static} stereo warping 
with the 
rectified baseline and \textit{temporal} warping using the predicted 
pose. In 
this way, the temporal information is incorporated into the training of 
depth, which leads to more accurate estimation. To 
deal with the inconsistent illumination between the training image 
pairs, 
our network predicts the \textit{brightness 
transformation parameters} which align the brightness of source and 
target images during training on the fly. The evaluation on the EuRoC 
MAV dataset shows that the proposed brightness transformation 
significantly improves the depth estimation accuracy. 
To integrate 
the deep depth 
into VO system, we firstly initialize every new 3D 
point with the predicted depth with a metric scale. Then we adopt the 
\textit{virtual stereo term} proposed in Deep Virtual Stereo Odometry 
(DVSO)~\cite{yang2018deep} to incorporate the predicted pose into the 
non-linear optimization. Unlike DVSO which uses a semi-supervised 
monocular depth estimation network relying on auxiliary depth extracted 
from state-of-the-art stereo VO system~\cite{wang2017stereoDSO}, 
our network uses only stereo videos without any external depth 
supervision.

Although the illumination change is explicitly modeled, it is not the 
only factor which may violate the 
brightness constancy assumption~\cite{klodt2018supervising}. Other 
factors, e.g., non-Lambertian surfaces, high-frequency areas and moving 
objects, also corrupt it. 
Inspired by the recent research on 
aleatoric uncertainty by deep neural 
networks~\cite{kendall2017uncertainties,klodt2018supervising}, the 
proposed network estimates the photometric uncertainty as predictive 
variance conditioned on the input image. As a result, the 
errors originated from pixels which are likely to violate the 
brightness constancy assumption are down-weighted. The learned weights 
of the photometric residuals also drive us to the idea of incorporating 
it into direct VO -- since both the self-supervised 
training scheme and the direct VO share a similar photometric 
objective, we propose to use the learned weights to replace the 
weighting function of the photometric residual in traditional direct VO 
which is empirically set~\cite{schops2019bad} or only accounts for the 
intrinsic uncertainty of the specific algorithm 
itself~\cite{kerl2013dense,engel2017direct}.

Robustness is one of the most important factors in 
designing VO 
algorithm. However, traditional monocular visual VO suffers from a lack 
of robustness 
when confronted with low textured areas or fast 
movement~\cite{von2018direct}. The typical 
solution is to introduce an inertial measurement unit (IMU). But this
increases the calibration effort and, more importantly, at constant 
velocity, IMUs cannot deliver the metric scale 
in constant velocity~\cite{martinelli2014closed}. We propose to increase the 
robustness of monocular VO by incorporating the estimated pose from 
the 
deep network into both the front-end tracking and the back-end 
non-linear 
optimization. For the front-end 
tracking, we replace the pose from the constant velocity motion model 
with 
the estimated pose from the network. Besides, the estimated pose is 
also used as a 
squared 
regularizer in addition to direct image 
alignment~\cite{szeliski2006image}. For the 
back-end non-linear optimization, we propose a pose energy term which 
is jointly minimized with the photometric energy term of direct VO. 

We evaluate the proposed monocular depth estimation network and D3VO 
on both KITTI~\cite{Geiger2012CVPR} and EuRoC MAV~\cite{Burri25012016}. We 
achieve state-of-the-art performances on both 
monocular depth estimation and camera tracking. In 
particular, by incorporating deep depth, deep 
uncertainty and deep pose, D3VO achieves comparable results to 
state-of-the-art stereo/LiDAR methods 
on KITTI Odometry, and also comparable results to 
the state-of-the-art VIO methods on EuRoC MAV, while being a 
monocular method.

\section{Related Work}
\textbf{Deep learning for monocular depth estimation.} 
Supervised learning~\cite{eigen2014depth,li2015depth,laina2016deeper} 
shows 
great performance on monocular depth estimation. Eigen et 
al.~\cite{eigen2014depth,eigen2015predicting} propose to use 
multi-scale CNNs 
which directly regresses the pixel-wise depth map from a single input 
image. 
Laina et al.~\cite{laina2016deeper} propose a robust loss function to 
improve 
the estimation accuracy. Fu et al.~\cite{fu:hal-01741163} recast the 
monocular 
depth estimation network as an ordinal regression problem and achieve 
superior performance. More recent works start to tackle the problem in a
self- and unsupervised way by learning the depth map using the photometric 
error~\cite{godard2016unsupervised,zhou2017unsupervised, 
yin2018geonet,mahjourian,zhan2018un,wang2018learning,Gordon_2019_ICCV} and 
adopting differentiable interpolation~\cite{jaderberg2015spatial}. Our 
self-supervised depth estimation network builds upon 
MonoDepth2~\cite{Godard_2019_ICCV} and extends it by predicting 
the brightness transformation parameters and the photometric uncertainty.  

\textbf{Deep learning for uncertainty estimation.} The 
uncertainty estimation of deep learning has recently been investigated in 
~\cite{kendall2017uncertainties,kendall2017multi} where two types 
of uncertainties are proposed. Klodt et 
al.~\cite{klodt2018supervising} propose to leverage the concept of 
aleatoric uncertainty to estimate the photometric and the depth 
uncertainties in order to improve the depth estimation accuracy. However, 
when formulating the photometric uncertainty, they do not consider 
brightness changes across different images which in fact can be 
modeled explicitly. Our method predicts 
the photometric uncertainty conditioned on the brightness-aligned 
image, which can deliver better photometric uncertainty estimation. Besides, we 
also seek to make better use of our 
learned 
uncertainties and
propose to incorporate them into traditional VO systems~\cite{engel2017direct}. 

\textbf{Deep learning for VO / SLAM.}
End-to-end learned deep neural networks have been explored to directly 
predict the relative poses between images with 
supervised~\cite{ummenhofer2017demon,wang2017deepvo, zhou2018deeptam} or unsupervised 
learning~\cite{zhou2017unsupervised,li2017undeepvo,wang2018learning,zhan2018un}.
Besides pose estimation, CodeSLAM~\cite{bloesch2018codeslam} delivers
dense reconstruction by jointly optimizing the learned prior of the
dense geometry together with camera poses. However, in terms of pose
estimation accuracy all these end-to-end methods are inferior to
classical stereo or visual inertial based VO methods.  Building on the
success of deep monocular depth estimation, several works integrate
the predicted depth/disparity map into monocular VO
systems~\cite{tateno2017cnn,yang2018deep} to improve performance and
eliminate the scale drift.  CNN-SLAM~\cite{tateno2017cnn} fuses the
depth predicted by a supervised deep neural network into
LSD-SLAM~\cite{engel2014lsd} and the depth maps are refined with
Bayesian filtering, achieving superior performance in indoor
environments~\cite{handa2014benchmark,sturm2012benchmark}. Other 
works~\cite{tang2019gcnv2,detone2018self} explore the application of deep 
neural networks on feature based methods ,and~\cite{jung2019corl} uses 
Generative 
Adversarial Networks (GANs) as an image enhancement method to improve the 
robustness of VO  in low light. The most
related work to ours is Deep Virtual Stereo Odometry (DVSO). DVSO
proposes a virtual stereo term that incooperates the depth estimation
from a semi-supervised network into a direct VO pipeline. In
particular, DVSO outperforms other monocular VO systems by a large
margin, and even achieves comparable performance to state-of-the-art
stereo visual odometry
systems~\cite{wang2017stereoDSO,mur2017orb}. While DVSO merely
leverages the depth, the proposed D3VO exploits the power of deep
networks on multiple levels thereby incorporating more information
into the direct VO pipeline.

\section{Method}
We first introduce a novel self-supervised neural network that
predicts depth, pose and uncertainty. The network also estimates
\textit{affine brightness transformation parameters} to align the
illumination of the training images in a self-supervised manner. The
photometric uncertainty is predicted based on a distribution over the
possible brightness
values~\cite{klodt2018supervising,kendall2017uncertainties} for each
pixel.  Thereafter we introduce D3VO as a direct visual odometry
framework that incorporates the predicted properties into both the
tracking front-end and the photometric bundle adjustment backend.
\begin{figure}
	\centering
	\includegraphics[width=\linewidth]{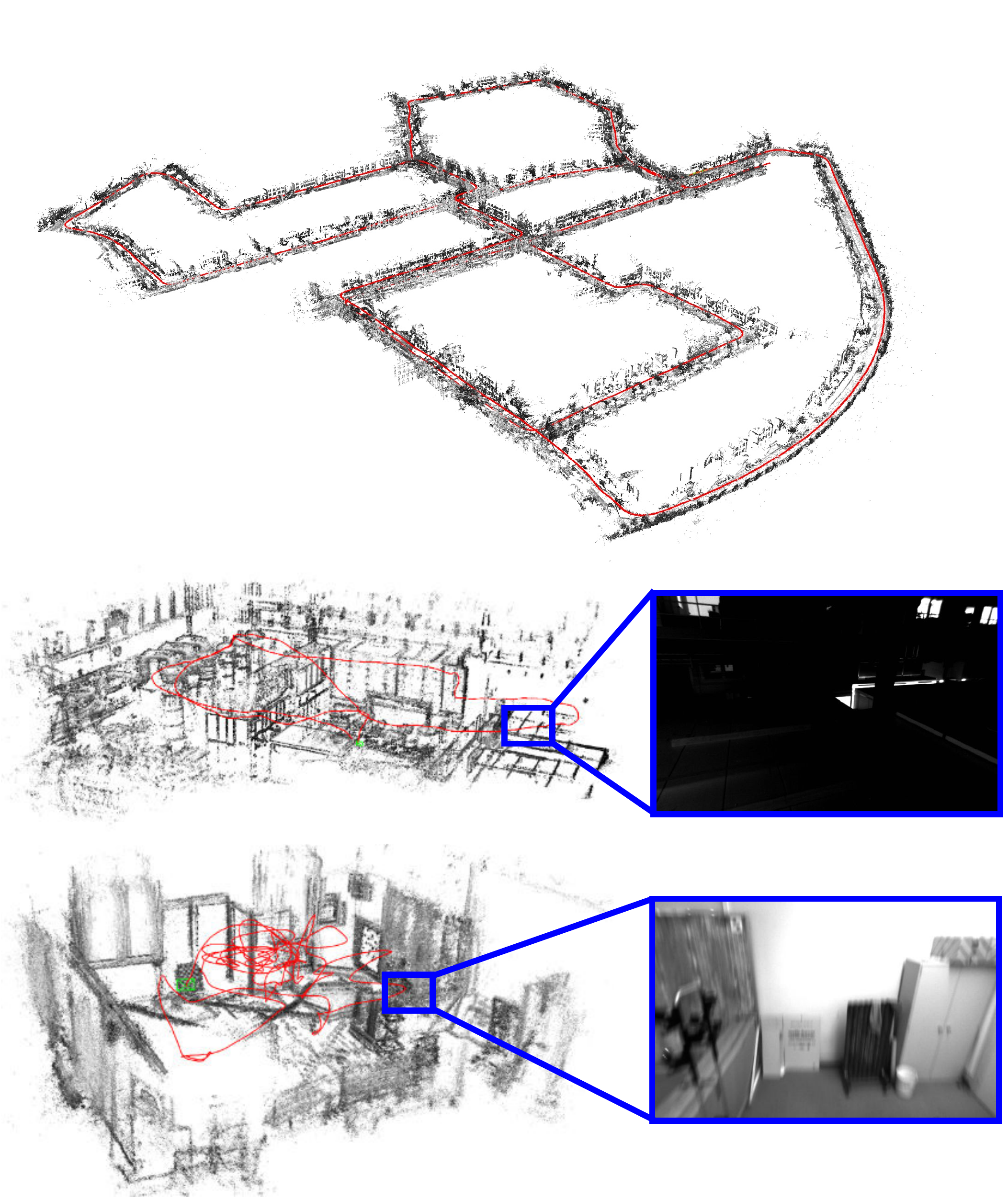}
	\caption{Examples of point clouds and trajectories delivered by D3VO on 
	KITTI Odometry Seq. 00, EuRoC 
	\textit{MH\_05\_difficult} and \textit{V1\_03\_difficult}. The insets on 
	EuRoC show the 
	scenarios with low illumination and motion blur which are among the main 
	reasons 
	causing failures of traditional purely vision-based VO systems.}\vspace{-1.5em}
	\label{fig:pc_exp}
\end{figure}

\subsection{Self-supervised Network}
The core concept of the proposed monocular depth estimation network is the 
self-supervised training scheme which simultaneously learns 
depth with DepthNet and 
motion with PoseNet using video 
sequences~\cite{zhou2017unsupervised,Godard_2019_ICCV}. The 
self-supervised training is realized by minimizing the minimum 
of the photometric re-projection errors between the temporal 
and static stereo images:
\begin{equation}
L_{self} = \frac{1}{|V|}\sum_{\mathbf{p}\in V}\min_{t'} 
r(I_t, 
I_{t'\rightarrow t}).
\label{eq:selfsup_min}
\end{equation}
where $V$ is the set of all pixels on $I_t$ and $t'$ is the 
index of all source frames. In our setting $I_t$ is the left 
image and $I_{t'}$ 
contains its two adjacent temporal frames and its 
opposite (right) frame, 
i.e., $I_{t'} \in \{I_{t-1}, I_{t+1}, I_{t^s}\}$. The per-pixel 
minimum loss is proposed in Monodepth2~\cite{Godard_2019_ICCV} 
in order to handle the occlusion among different source frames.
To simplify notation, we use $I$ 
instead of $I(\mathbf{p})$ in the remainder of this 
section. $I_{t'\rightarrow t}$ is 
the sythesized $I_t$ by warping the temporal stereo images with 
the predicted depth $D_t$, the camera pose $\mathbf{T}_t^{t'}$, the 
camera intrinsics 
$K$, and the differentialble bilinear 
sampler~\cite{jaderberg2015spatial}. 
Note that for $I_{t^s\rightarrow t}$, the transformation 
$\mathbf{T}_t^{t^s}$ is known and constant. DepthNet also predicts the 
depth map $D_{t^s}$ of the right image $I_{t^s}$ by feeding 
only the left image $I_t$ as proposed 
in~\cite{godard2016unsupervised}. The training of $D_{t^s}$ 
requires to synthesize $I_{t\rightarrow {t^s}}$ and compare 
with $I_{t^s}$. For simplicity, we will in  the following only detail the loss 
regarding the left image.

The common 
practice~\cite{godard2016unsupervised} is to formulate the 
photometric error as
\begin{equation}
r(I_a, I_b) = \frac{\alpha}{2}(1 - \text{SSIM}(I_a, I_b)) + (1-\alpha)||I_a 
- I_b||_1
\label{eq:residual}
\end{equation}
based on the brightness constancy 
assumption. However, it can be violated due to illumination changes and 
auto-exposure of the camera to which both 
L1 and SSIM~\cite{wang2004image} are not invariant. 
Therefore, we propose to explicitly model the camera exposure change with 
predictive \textit{brightness transformation 
parameters}.

\begin{figure}
	\centering
	\includegraphics[width=\linewidth]{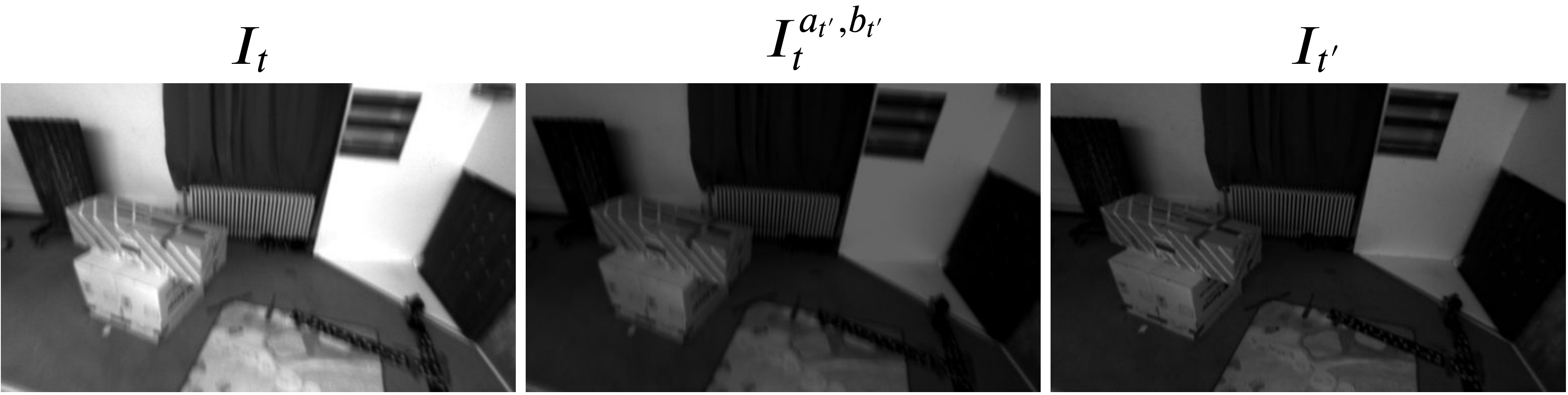}
	\caption{Examples of affine brightness transformation on 
	EuRoC MAV~\cite{Burri25012016}. Originally the source 
		image ($I_{t'}$) and the target image ($I_t$) show 
		different brightness. With the predicted 
		parameters $a,b$, the transformed target images 
		$I^{a',b'}$ have similar brightness as the source 
		images, which facilitates the self-supervised 
		training based on the brightness constancy assumption.
	}\vspace{-1.5em}
	\label{fig:affine}
\end{figure}

\textbf{Brightness transformation parameters}. The change of 
the image intensity due to 
the adjustment of camera exposure can be modeled as an affine 
transformation 
with two parameters $a, b$
\begin{equation}
	I^{a,b} = aI + b.
\end{equation}
Despite its simplicity, this formulation has been shown 
to be effective in direct 
VO/SLAM, 
e.g.,~\cite{engel2015large,engel2017direct,wang2017stereoDSO,jin2001real}, 
which builds upon the brightness constancy assumption as well. 
Inspired by 
these works, we 
propose predicting the transformation parameters $a,b$ which 
align the brightness condition of $I_t$ with $I_{t'}$. We reformulate 
Eq.~\eqref{eq:selfsup_min} as
\begin{equation}
	L_{self} = \frac{1}{|V|}\sum_{\mathbf{p}\in V}\min_{t'} 
	r(I_t^{a_{t'},b_{t'}}, 
	I_{t'\rightarrow t})
\label{eq:loss_with_affine}
\end{equation}
with
\begin{equation}
	I_t^{a_{t'},b_{t'}} = a_{t \rightarrow t'}I_t + b_{t 
	\rightarrow t'},
\end{equation}
where $a_{t \rightarrow t'}$ and $b_{t \rightarrow t'}$ are the 
transformation parameters aligning the illumination of $I_t$ to 
$I_{t'}$. Note that both 
parameters can be 
trained in a self-supervised way without any supervisional signal. 
Fig.~\ref{fig:affine} shows the affine transformation examples 
from EuRoC MAV~\cite{Burri25012016}.

\textbf{Photometric uncertainty}. Only modeling affine brightness 
change is not enough to capture all failure cases of 
the brightness constancy assumption. Other cases like 
non-Lambertian surfaces and moving objects, are 
caused by the intrinsic properties of the corresponding 
objects which are not trivial to model 
analytically~\cite{klodt2018supervising}. Since these 
aspects can be seen as 
observation noise, we leverage the concept of heteroscedastic 
aleatoric uncertainty of deep neural networks proposed by 
Kendall et 
al.~\cite{kendall2017uncertainties}. The key idea is to
predict a posterior probability distribution for each pixel 
parameterized with its mean as well as its variance $p(y|\tilde{y}, 
\sigma)$ over ground-truth labels $y$. For 
instance, by assuming 
the noise is Laplacian, the negative log-likelihood to be 
minimized is
\begin{equation}
	-\log p(y|\tilde{y}, \sigma) = \frac{|y - \tilde{y}|}{\sigma} + \log 
	\sigma + const.
	\label{eq:uncer}
\end{equation}
Note that no ground-truth label for $\sigma$ is needed for 
training. The predictive uncertainty allows the network to 
adapt the weighting of the residual dependent on the data 
input, which improves the 
robustness of the model to noisy data or erroneous 
labels~\cite{kendall2017uncertainties}.

In our case where the ``ground-truth'' $y$ are the pixel 
intensities on the target images, the network will predict 
higher $\sigma$ for the pixel areas on $I_t$ where the 
brightness constancy assumption may be 
violated. Similar to~\cite{klodt2018supervising}, 
we 
implement this by converting Eq.~\eqref{eq:loss_with_affine} to
\begin{equation}
	L_{self} = \frac{1}{|V|}\sum_{\mathbf{p}\in 
	V}\frac{\min_{t'} r(I_t^{a_{t'}, b_{t'}}, 
	I_{t'\rightarrow t})}{\Sigma_t} + \log \Sigma_t,
\end{equation}
where $\Sigma_t$ is the uncertainty map of $I_t$. 
Fig.~\ref{fig:kitti_depth} shows 
the qualitative results of the predicted uncertainty maps on 
KITTI~\cite{Geiger2012CVPR} and EuRoC~\cite{Burri25012016} 
datasets, respectively. In the next 
section, we will show that the learned $\Sigma_t$ is useful for weighting the 
photometric residuals for D3VO.

The total loss function is the summation of the self-supervised 
losses and the regularization losses on multi-scale images:
\begin{equation}
 L_{total} = \frac{1}{s}\sum_{s} (L_{self}^s + \lambda 
 L_{reg}^s),
\end{equation}
where $s=4$ is the number of scales and
\begin{equation}
L_{reg} = L_{smooth} + \beta L_{ab}
\end{equation}
with
\begin{equation}
	L_{ab} = \sum_{t'} (a_{t'} - 1)^2 + 
	b_{t'}^2
\end{equation}
is the regularizer of the brightness parameters and $L_{smooth}$ is the 
edge-aware smoothness on $D_t$~\cite{godard2016unsupervised}.

To summarize, the proposed DepthNet predicts $D_t$, $D_{t^s}$ 
and $\Sigma_t$ with one single input $I_t$. 
PoseNet predicts $\mathbf{T}_t^{t'}$, $a_{t\rightarrow t'}$ and 
$b_{t\rightarrow t'}$ with channel-wise concatenated ($I_t$, 
$I_{t'}$) as the input. Both DepthNet and PoseNet are 
convolutional networks following the widely used UNet-like 
architecture~\cite{ronneberger2015u}. Please refer to our 
supplementary materials for network architecture and implementation 
details.

\subsection{D3VO}
In the previous section, we introduced the self-supervised 
depth estimation 
network which predicts the depth map $D$, the 
uncertainty map $\Sigma$ and the relative pose
$\mathbf{T}_t^{t'}$. In this section, we will describe how 
D3VO 
integrates these 
predictions into a windowed sparse photometric bundle adjustment 
formulation as proposed in~\cite{engel2017direct}. Note 
that in the following we use $\ \widetilde{\cdot}\ $ denoting 
the predictions from the network as $\widetilde{D}$, 
$\widetilde{\Sigma}$ and $\widetilde{\mathbf{T}}_t^{t'}$ to avoid 
ambiguity.

\textbf{Photometric energy}. D3VO 
aims to minimize a 
total photometric error $E_{photo}$ defined as

\begin{equation}
	E_{photo} = \sum_{i\in \mathcal{F}}\sum_{\mathbf{p} \in 
	\mathcal{P}_i}\sum_{j\in\text{obs}(\mathbf{p})}E_{\mathbf{p}j},
	\label{eq:pe}
\end{equation}
where $\mathcal{F}$ is the set of all keyframes, 
$\mathcal{P}_i$ is the set of 
points hosted in keyframe $i$, obs($\mathbf{p}$)  is the set of 
keyframes 
in which point $\mathbf{p}$ is observable and $E_{\mathbf{p}j}$ is 
the weighted photometric energy term when $\mathbf{p}$ is projected onto keyframe $j$:
\begin{equation}
	E_{\mathbf{p}j} := 
	\sum_{\mathbf{p}\in\mathcal{N}_\mathbf{p}}w_\mathbf{p}\left|
	 \left| 
	(I_j[\mathbf{p'}] - b_j) - 
	\frac{e^{a_j}}{e^{a_i}}(I_i[\mathbf{p}] - 
	b_i) \right| \right|_\gamma,
\end{equation}
where $\mathcal{N}$ is the set of 8 neighboring pixels of 
$\mathbf{p}$ defined in~\cite{engel2017direct}, $a$,$b$ are the 
affine 
brightness parameters jointly estimated by non-linear optimization as 
in~\cite{engel2017direct} and 
$||\cdot||_\gamma$ is the 
Huber norm. In~\cite{engel2017direct}, the residual
is down-weighted when the pixels are with high image gradient to 
compensate small independent geometric 
noise~\cite{engel2017direct}. In realistic 
scenarios, there are 
more sources of noise, e.g.,
reflection\cite{klodt2018supervising}, that need to be 
modeled in order to deliver accurate and robust motion 
estimation. We propose 
to use the learned uncertainty $\widetilde{\Sigma}$ to 
formulate the 
weighting 
function
\begin{equation}
w_\mathbf{p} = \frac{\alpha^2}{\alpha^2 + \left| 
	\left|\widetilde{\Sigma}(\mathbf{p})\right| 
	\right|_2^2},
\end{equation}
which may not only 
depend on local image gradient, but also on higher level noise 
pattern. As shown in Fig.~\ref{fig:kitti_depth}, the proposed network is able 
to predict high uncertainty on 
the areas of reflectance, e.g., the windows of the vehicles, the moving object 
like the cyclist and the object boundaries where depth discontinuity occurs.  

The projected point position of $\mathbf{p'}$ is given by
$
	\mathbf{p'} = \Pi(\mathbf{T}_i^j\Pi^{-1}(\mathbf{p}, 
	d_{\mathbf{p}})),
$
where $d_{\textbf{p}}$ is the depth of the point $\mathbf{p}$ 
in the coordinate 
system of keyframe $i$ and $\Pi(\cdot)$ is the projection 
function 
with the known 
camera intrinsics.
Instead of randomly initializing $d_{\mathbf{p}}$ 
as in 
traditional monocular direct 
methods~\cite{engel2014lsd,engel2017direct}, we 
initialize the point with $d_{\mathbf{p}} = \widetilde{D}_i[\mathbf{p}]$ 
which provides the metric 
scale. Inspired by~\cite{yang2018deep}, we introduce a 
\textit{virtual stereo term} $E_{\mathbf{p}}^{\dagger}$ to 
Eq.~\eqref{eq:pe}
\begin{equation}
	E_{photo} = \sum_{i\in \mathcal{F}}\sum_{\mathbf{p} \in 
		\mathcal{P}_i}\left(\lambda E_{\mathbf{p}}^{\dagger} + 
		\sum_{j\in\text{obs}(\mathbf{p})}E_{\mathbf{p}j}\right)
\end{equation}
with
\begin{equation}
E_{\mathbf{p}}^{\dagger} = w_{\mathbf{p}}\left|\left| 
I_i^\dagger[\mathbf{p^\dagger}] - 
I_i[\mathbf{p}]\right|\right|_\gamma,
\end{equation}
\begin{equation}
	I_i^\dagger[\mathbf{p^\dagger}] = 
	I_i[\Pi(\mathbf{\mathbf{T}_s}^{-1}\Pi^{-1}(\mathbf{p}^\dagger, 
	D_{i^s}[\mathbf{p}^\dagger]))]
\end{equation}
with $\mathbf{\mathbf{T}_s}$ the transformation matrix from the left to the 
right image used for training DepthNet and
\begin{equation}
	\mathbf{p}^\dagger = 
	\Pi(\mathbf{\mathbf{T}_s}\Pi^{-1}(\mathbf{p}, d_{\mathbf{p}})).
\end{equation}
The virtual stereo term optimizes the estimated depth 
$d_{\mathbf{p}}$ from VO to be consistent with the depth 
predicted by the proposed deep network~\cite{yang2018deep}.

\textbf{Pose energy}. Unlike traditional direct 
VO approaches~\cite{engel2013iccv,forster2014svo} which initialize the 
front-end tracking for each new frame with a constant velocity motion model, we 
leverage the predicted poses between consecutive frames to build a non-linear 
factor graph~\cite{kschischang2001factor,loeliger2004introduction}. 
Specifically, we create a new factor graph whenever the newest keyframe, which 
is also the reference frame for the front-end tracking, is updated. Every new 
frame is tracked with respect to the reference keyframe with direct image 
alignment~\cite{szeliski2006image}. Additionally, the predicted relative pose 
from the deep network is used as a factor between the current frame and the 
last frame. After the optimization is finished, we marginalize the last frame 
and the factor graph will be used for the front-end tracking of the following 
frame. Please refer to our supp. materials for the visualization of the factor 
graph.

The pose estimated from the tracking front-end is then used to initialize the 
photometric
bundle adjustment backend. We further introduce a prior for the relative keyframe pose
$\mathbf{T}_{i-1}^{i}$ using the predicted pose $\widetilde{\mathbf{T}}_{i-1}^{i}$.
Note that $\widetilde{\mathbf{T}}_{i-1}^{i}$ is calculated 
by concatenating all the predicted frame-to-frame poses
between keyframe $i-1$ and $i$. Let

\begin{equation}
E_{pose} = \sum_{i \in \mathcal{F} - 
	\{0\}}\text{Log}(\widetilde{\mathbf{T}}_{i-1}^{i}
\mathbf{T}^{i-1}_{i})^{\top 
	}\bm{\Sigma}_{\widetilde{\bm{\xi}}_{i-1}^{i}}^{-1}\text{Log}(\widetilde{\mathbf{T}}_{i-1}^{i}
	\mathbf{T}^{i-1}_{i}),
\label{eq:pose_err2}
\end{equation}
where $\text{Log: SE(3)}\rightarrow \mathbb{R}^6$
maps from the transformation matrix $\mathbf{T} \in 
\mathbb{R}^{4 \times 4}$ in the Lie group $\text{SE}(3)$ to 
its corresponding twist coordinate $\bm{\xi} \in 
\mathbb{R}^6$ in the Lie algebra $\mathfrak{se}(3)$.
The diagonal inverse covariance matrix 
$\bm{\Sigma}_{\widetilde{\bm{\xi}}_{i-1}^{i}}^{-1}$ is obtained by
propagating the covariance matrix between each consecutive frame pairs that is modeled
as a constant diagonal matrix.

The total energy function is defined as
\begin{equation}
	E_{total} = E_{photo} + w E_{pose}.
	\label{eq:e_total}
\end{equation}

\begin{table*}[tp]
	\centering
	\small
	\begin{tabular}{lcccccccccc}
		&    & &RMSE   & RMSE (log)   & ARD  & SRD  &  & 
		$\delta < 1.25$ & 
		$\delta < 1.25^2$ & $\delta < 1.25^3$ \\ \cline{4-7} 
		\cline{9-11} 
		Approach & Train &  & \multicolumn{4}{c}{lower is 
			better} &  & 
		\multicolumn{3}{c}{higher is better} \\
		\hline
		MonoDepth2\cite{godard2016unsupervised}  & 
		MS&& 
		4.750 & 0.196 & 0.106 & 0.818 && 
		0.874 & \textit{0.957} & 0.979 \\
		Ours, \textit{uncer}
		&MS&&
		\textit{4.532} & \textit{0.190} & 
		\textit{0.101} & \textit{0.772} && 
		\textit{0.884} & 0.956 & \textit{0.978} \\
		Ours, \textit{ab}
		&MS&&
		4.650 & 0.193 & 
		0.105 & 0.791 && 
		0.878 & \textit{0.957} & 0.979 \\
		Ours, \textit{full}
		&MS&&
		\textbf{4.485} & \textbf{0.185} & 
		\textbf{0.099} & \textbf{0.763} && 
		\textbf{0.885} & \textbf{0.958} & \textbf{0.979} \\
		\hline
		\hline
		Kuznietsov et al.~\cite{kuznietsov2017semi} &DS&& 
		4.621 & 0.189 & 
		0.113 & \textit{0.741} && 
		0.862 & \textbf{0.960} & \textbf{0.986} \\
		DVSO~\cite{yang2018deep} 
		&D*S&&
		\textbf{4.442} & \textit{0.187} & 
		\textbf{0.097} & \textbf{0.734} && 
		\textbf{0.888} & 0.958 & \textit{0.980} \\
		Ours 
		&MS&&
		\textit{4.485} & \textbf{0.185} & 
		\textit{0.099} & 0.763 && 
		\textit{0.885} & 0.958 & 0.979 \\
		\hline
	\end{tabular}
	\vspace{-0.8em}
	\caption{
		Depth evaluation results on the KITTI 
		Eigen split~\cite{eigen2014depth}. M: self-supervised 
		monocular supervision; S: self-supervised stereo supervision;
		D: ground-truth depth supervison; D*: sparse auxiliary depth 
		supervision. 
		The upper part shows the comparison with the SOTA 
		self-supervised network Monodepth2~\cite{Godard_2019_ICCV} under the 
		same setting and the 
		ablation study of the brightness 
		transformation parameters (\textit{ab}) and the photometric uncertainty 
		(\textit{uncer}).
		The lower part shows the comparison with the SOTA 
		\textit{semi}-supervised methods using stereo as well as depth 
		supervision. Our method outperforms Monodepth2 on all metrics and can 
		also deliver 
		comparable performance to the SOTA semi-supervised method
		DVSO~\cite{yang2018deep} that additionally uses 
		the depth from Stereo DSO~\cite{wang2017stereoDSO} as sparse 
		supervision signal.}\vspace{-1em}
	\label{tab:kitti_depth}
\end{table*}

Including the pose prior term $E_{pose}$ in Eq.~\ref{eq:e_total} can be considered as an analogy to 
integrating the pre-integrated IMU pose prior into the system with a Gaussian 
noise model. 
$E_{total}$ is minimized using the Gauss-Newton method. To summarize, we boost 
the direct VO method by introducing the predicted poses as initializations to both the tracking 
front-end and the optimization backend, as well as adding them as a regularizer 
to the energy function of the photometric bundle 
adjustment.

\section{Experiments}
We evaluate the proposed self-supervised monocular depth 
estimation network as well as 
D3VO on both the KITTI~\cite{Geiger2012CVPR} and the EuRoC 
MAV~\cite{Burri25012016} datasets.

\subsection{Monocular Depth Estimation}
\textbf{KITTI}. We train and evalutate the proposed 
self-supervised depth 
estimation 
network on the split of Eigen at el.~\cite{eigen2014depth}. The 
network is 
trained on 
stereo sequences with the pre-processing proposed by Zhou et 
al.~\cite{zhou2017unsupervised}, which gives us 39,810 training 
quadruplets, each of which 
contains 3 (left) temporal images and 1 (right) stereo image, 
and 4,424 for validation. 
The upper part of Table~\ref{tab:kitti_depth} shows the 
comparison with Monodepth2~\cite{Godard_2019_ICCV} 
which is the state-of-the-art method trained with stereo and 
monocular setting, and also the ablation study of the 
proposed brightness transformation prediction (\textit{ab}) and the photometric 
uncertainty estimation (\textit{uncer}). The results demonstrate that the proposed depth 
estimation network outperforms Monodepth2 on all metrics. 
The ablation studies unveil that the significant improvement over Monodepth2 
comes largely with \textit{uncer}, possibly because in KITTI there are 
many objects with non-Lambertian surfaces like windows and also 
objects that move independently such as cars and leaves which 
violate the brightness constancy assumption.
The lower part of the table shows the comparison to the 
state-of-the-art \textit{semi}-supervised methods and the results show 
that our method can achieve competitive performance 
without using any depth supervision.

\begin{figure}
	\centering
	\includegraphics[width=\linewidth]{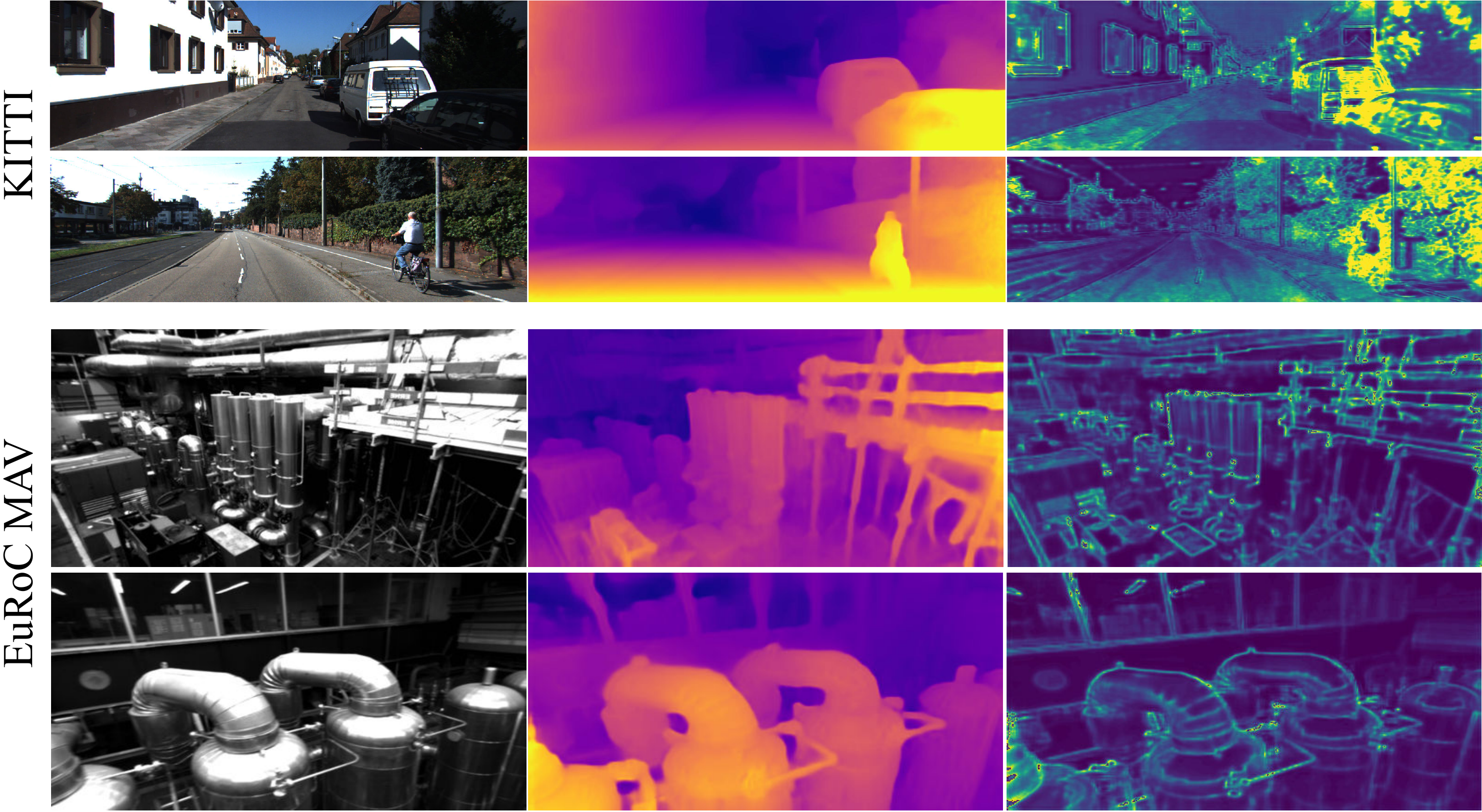}
	\caption{Qualitative results from KITTI and EuRoC MAV. The original 
	image, the predicted depth maps and the uncertainty maps are shown from the left to the right, 
	respectively. In particular, the network is able to predict 
	high uncertainty on object boundaries, moving objects, highly reflecting 
	and high frequency areas.}
\vspace{-1em}
	\label{fig:kitti_depth}
\end{figure}

In Figure~\ref{fig:kitti_depth} we show some qualitative 
results obtained from the Eigen test 
set~\cite{eigen2014depth}. 
From left to right, the original image, the depth maps and the uncertainty maps
are shown respectively.
For more qualitative results and the generalization capability on the 
Cityscapses dataset~\cite{cordts2016cityscapes}, please refer 
to our supp. materials. 

\begin{table}
	\centering
	\small
	\setlength{\tabcolsep}{0.5em}
	\begin{tabular}{lccccc}
		\hline
		&    RMSE   & RMSE (log)   & ARD  & SRD  & $\delta < 
		1.25$ \\ 
		\hline
		Monodepth2
		&
		0.370 & 0.148 & 
		0.102 & 0.065 &
		0.890 \\
		Ours, \textit{ab}
		&
		\textit{0.339} & \textit{0.130} & 
		\textit{0.086} & \textit{0.054} &
		\textit{0.929} \\
		Ours, \textit{uncer}
		&
		0.368 & 0.144 & 
		0.100 & 0.065 &
		0.892 \\
		Ours, \textit{full} 
		&
		\textbf{0.337} & \textbf{0.128} & 
		\textbf{0.082} & \textbf{0.051} &
		\textbf{0.931} \\
		\hline
	\end{tabular}
	\vspace{-1em}
	\caption{Evaluation results of \textit{V2\_01} in EuRoC MAV~\cite{Burri25012016}. 
		The performance of monocular depth estimation is boosted 
		largely by the proposed predictive brightness transformation parameters.
	}
	\label{tab:euroc_our_split}
\end{table}

\begin{table}
	\centering
	\small
	\begin{tabular}{lccccc}
		\hline
		&    RMSE   & RMSE (log)   & ARD  & SRD  & $\delta < 
		1.25$ \\ 
		\hline
		\cite{Gordon_2019_ICCV} 
		&
		0.971 & 0.396 & 
		0.332 & 0.389 &
		0.420 \\
		Ours
		&
		\textbf{0.943} & \textbf{0.391} & 
		\textbf{0.330} & \textbf{0.375} &
		\textbf{0.438} \\
		\hline
	\end{tabular}
	\vspace{-1em}
	\caption{Evaluation results of \textit{V2\_01} in EuRoC MAV~\cite{Burri25012016} 
		with the model trained with all \textit{MH} sequences. 
	}
	\label{tab:google}
	\vspace{-1em}
\end{table}

\textbf{EuRoC MAV}. The EuRoC MAV 
Dataset~\cite{Burri25012016} is a dataset containing 11 sequences 
categorized as \textit{easy}, \textit{medium} and \textit{difficult} according to 
the illumination and camera motion. 
This dataset is very challenging due to the strong motion and 
significant illumination 
changes both between stereo and temporal images. 
We therefore consider it as a nice test bench for validating the effectiveness of our
predictive brightness transformation parameters for depth prediction.
Inspired by Gordon et al.~\cite{Gordon_2019_ICCV} who recently generated ground truth
depth maps for the sequence \textit{V2\_01} by projecting the provided Vicon 3D scans and filtering
out occluded points, we also use this sequence for depth evaluations\footnote{We thank the 
authors of~\cite{Gordon_2019_ICCV} to provide the processing code.}. Our first 
experiment is set up to be 
consistent
as in~\cite{Gordon_2019_ICCV}, for which we train models with the monocular 
setting on all \textit{MH} sequences 
and test on 
\textit{V2\_01} and show the results in Table~\ref{tab:google}.

In the second experiment, we use 5 sequences \textit{MH\_01}, \textit{MH\_02}, 
\textit{MH\_04}, \textit{V1\_01} and \textit{V1\_02} as the training set to check the performance 
of our method in a relatively loosened setting. We remove the static frames for training and 
this results in 12,691 images of which 11,422 images are used for training and 1269 images are used 
for validation.
We train our model with different ablations, as well as 
Monodepth2~\cite{Godard_2019_ICCV} as the 
baseline. 
The results in Table~\ref{tab:euroc_our_split} show that all our variations outperform the baseline 
and, in contrast to the case in KITTI, the proposed \textit{ab} improves the 
results on this 
dataset significantly. Please refer to the supp. materials for more 
experiments 
on \textit{ab}. In fact, it 
is 
worth noting that the results in 
Table~\ref{tab:google} (trained on one scene \textit{MH} and 
tested on another scene \textit{V}) are worse than the ones in 
Table~\ref{tab:euroc_our_split} 
(trained on both \textit{MH} and \textit{V}), 
which implies that it is still a challenge to improve the generalization 
capability of monocular depth estimation among very different scenarios.

\subsection{Monocular Visual Odometry}
We evaluate the VO performance of D3VO on both KITTI Odometry and 
EuRoC MAV with the network trained on the splits described in the previous 
section.

\textbf{KITTI Odometry}. The KITTI Odometry Benchmark 
contains 11 (0-10) 
sequences with provided ground-truth poses. As summarized 
in~\cite{yang2018deep}, 
sequences 00, 03, 04, 05, 07 are in the training set of the Eigen split
that the proposed network uses, so we consider the rest of the sequences as the 
testing set for evaluating the pose estimation of D3VO. We use the relative 
translational ($t_{rel}$) error proposed 
in~\cite{Geiger2012CVPR} as the main metric for evaluation.  
Table~\ref{tab:kitti_vo_eval} shows the 
comparison with other state-of-the-art \textit{mono} (M) as well as 
\textit{stereo} (S) VO 
methods on the rest 
of the sequences. We refer to~\cite{yang2018deep} for the results of the 
compared methods. Traditional monocular methods show high errors in the 
large-scale outdoor scene like the sequences in KITTI due to the scale drift.  
D3VO achieves the best performance on average, despite being a monocular 
methods as well. The table also contains the
ablation study on the integration of deep depth (\textit{Dd}), pose 
(\textit{Dp}) and uncertainty (\textit{Du}). It can be noticed that, consistent 
with the results in Table~\ref{tab:kitti_depth}, the predicted uncertainty 
helps a lot on KITTI. 
We also submit the results on the testing sequences (11-20) to the 
KITTI Odometry evaluation  
server (\href{http://www.cvlibs.net/datasets/kitti/eval_odometry.php}{link}).
At the time of submission, D3VO outperforms DVSO and achieves the best 
monocular VO performance 
and comparable to other state-of-the-art LiDAR and stereo methods.

We further compare D3VO with state-of-the-art end-to-end deep learning 
methods and other recent hybrid methods and show the results in Table~\ref{tab:compare_deep}. Note 
that here we only show the results on Seq.09 and 10, since most of the 
end-to-end 
methods only provide the results on these two sequences. We refer 
to~\cite{Gordon_2019_ICCV,zhan2019visual,yang2018deep} for the results for the 
compared methods. D3VO 
achieves better performance than all the end-to-end methods by a 
notable margin. In general, hybrid methods which combine deep learning with 
traditional methods deliver better results than end-to-end methods.

\newcolumntype{L}[1]{>{\let\newline\\
		\arraybackslash\hspace{0pt}}m{#1}}
\newcolumntype{C}[1]{>{\centering\let\newline\\
		\arraybackslash\hspace{0pt}}m{#1}}
\begin{table}[]
	\small
	\setlength{\tabcolsep}{0.4em}
	\begin{tabular}{C{0.3cm}L{1.54cm}|C{0.6cm}C{0.6cm}C{0.6cm}C{0.6cm}C{0.6cm}C{0.6cm}|c}
		&& 01            & 02            & 06            & 
		08            & 09            & 10            & 
		mean          \\ \hline
		\parbox[t]{2mm}{\ 
			\multirow{2}{1mm}{\rotatebox[origin=c]{90}{\underline{\ \ M\ 
						\ }}}}&DSO~\cite{engel2017direct}
		& 9.17         & 
		114          & 
		42.2          & 177          & 28.1          & 
		24.0          & 65.8          \\
		&ORB~\cite{mur2015orb}     & 108         & 
		10.3         & 
		14.6          & 11.5          & 9.30          & 
		2.57          & 37.0          \\ \hline\hline
		\parbox[t]{2mm}{\ 
			\multirow{3}{1mm}{\rotatebox[origin=c]{90}{\underline{\ 
						\ S\ \ }}}}&S. LSD~\cite{engel2015large}     & 
		2.13           
		&1.09          & 
		1.28          & 1.24          & 1.22          & 
		0.75          & 1.29          \\
		&ORB2~\cite{mur2017orb}      & 1.38          & 
		0.81          & 
		0.82          & 1.07          & 0.82          & 
		\textit{0.58} & 0.91          \\
		&S. DSO~\cite{wang2017stereoDSO}       & 1.43          
		& \textbf{0.78} & 
		\textit{0.67} & \textbf{0.98} & 0.98          & 
		\textbf{0.49} & 0.89          \\ \hline
		&\textit{Dd}    & 1.16          & 
		0.84          & 
		0.71          & 1.01          & 0.82          & 
		0.73          & 0.88          \\
		&\textit{Dd+Dp} & 1.15          & 0.84          & 
		0.70          & 1.03          & \textit{0.80} & 
		0.72          & 0.87          \\
		&\textit{Dd+Du} & \textit{1.10} & 0.81          & 
		\textit{0.69} & 1.03          & \textbf{0.78} & 
		\textit{0.62}          & \textit{0.84} \\
		&D3VO         & \textbf{1.07} & \textit{0.80} & 
		\textbf{0.67} & \textit{1.00} & \textit{0.78} & 
		\textit{0.62} & \textbf{0.82}\\
	\end{tabular}
	\vspace{-1em}
	\caption{Results on our test split of KITTI 
		Odometry. The results of the SOTA 
		monocular (M) methods are shown as baselines. The comparison 
		with 
		the SOTA stereo (S) methods shows that D3VO achieves 
		better average 
		performance than other methods, while 
		being a monocular VO. We also show the ablation study 
		for 
		the integration 
		of deep depth(\textit{Dd}), pose(\textit{Dp}) as 
		well as uncertainty(\textit{Du}). }
	\label{tab:kitti_vo_eval}
	\vspace{-1em}
\end{table}

\newcolumntype{C}[1]{>{\centering\let\newline\\
		\arraybackslash\hspace{0pt}}m{#1}}
\begin{table}
	\centering
	\small
	\setlength{\tabcolsep}{0em}
	\begin{tabular}{cC{3.0cm}|C{2.0cm}|C{2.0cm}}
		&& \multicolumn{1}{c|}{Seq. 09} & 
		\multicolumn{1}{c}{Seq. 10}\\
		\hline
		\parbox[t]{2mm}{\ 
		\multirow{6}{1mm}{\rotatebox[origin=c]{90}{\underline{\ \ End-to-end\ \ 
		}}}}&
		\multicolumn{1}{c|}{UnDeepVO~\cite{li2017undeepvo}}
		& 7.01 & 10.63\\
		&\multicolumn{1}{c|}{SfMLearner~\cite{zhou2017unsupervised}}
		 &  17.84& 
		37.91\\
		&\multicolumn{1}{c|}{Zhan et al.~\cite{zhan2018un}}& 
		11.92& 12.45\\
		&\multicolumn{1}{c|}{Struct2Depth~\cite{casser2019depth}}&
 
		10.2& 28.9\\
		&\multicolumn{1}{c|}{Bian et 
		al.~\cite{bian2019depth}}&11.2 & 
		10.1\\
		&\multicolumn{1}{c|}{SGANVO~\cite{feng2019sganvo}}&\textit{4.95} & 
		\textbf{5.89}\\
		&\multicolumn{1}{c|}{Gordon et 
		al.~\cite{Gordon_2019_ICCV}} & 
		\textbf{2.7}& \textit{6.8}\\
		\hline
		\hline
		\parbox[t]{2mm}{\ 
			\multirow{5}{1mm}{\rotatebox[origin=c]{90}{\underline{\ \ Hybrid\ \ 
			}}}}&\multicolumn{1}{c|}{CNN-SVO~\cite{loo2019cnn}}
						 & 10.69 & 4.84\\
		&\multicolumn{1}{c|}{Yin et al.~\cite{yin2017scale}} & 
		4.14 & 1.70\\
		&\multicolumn{1}{c|}{Zhan et al.~\cite{zhan2019visual}} 
		& 2.61 & 2.29\\
		&\multicolumn{1}{c|}{DVSO~\cite{yang2018deep}} & 
		\textit{0.83} & 
		\textit{0.74}\\
		&\multicolumn{1}{c|}{D3VO} & \textbf{0.78} & 
		\textbf{0.62}\\
	\end{tabular}
	\vspace{-1em}
	\caption{Comparison to other hybrid methods as well as 
	end-to-end methods on 
	Seq.09 and 10 of KITTI Odometry.}
	\label{tab:compare_deep}
	\vspace{-1.5em}
\end{table}

\begin{table}
	\small
	\setlength{\tabcolsep}{0.4em}
\begin{tabular}{cl|ccccc|c}
	&& M03     & M05   & V103     & 
	V202     & V203 & mean          \\ \hline 
	\parbox[t]{2mm}{\ 
		\multirow{2}{1mm}{\rotatebox[origin=c]{90}{\underline{\ \ M\ \  
				}}}}&DSO~\cite{engel2017direct}&0.18&0.11&1.42&0.12&0.56&0.48\\
	&ORB~\cite{mur2015orb}&\textbf{0.08}&0.16&1.48&1.72&0.17&0.72\\
	\hline\hline
	\parbox[t]{2mm}{\ 
		\multirow{7}{1mm}{\rotatebox[origin=c]{90}{\underline{\ \ M+I\ 
				\ }}}}&VINS~\cite{qin2019general}  & 0.13          & 
	0.35          & 0.13          & 
	0.08          &0.21 & 0.18          \\
	&OKVIS~\cite{leutenegger2015keyframe} & 0.24          & 
	0.47          & 0.24          & 
	0.16          &0.29 & 0.28          \\
	&ROVIO~\cite{bloesch2015robust} & 0.25          & 
	0.52          & 0.14          & 
	0.14          &0.14 & 0.24         \\
	&MSCKF~\cite{mourikis2007multi} & 0.23          & 
	0.48          & 0.24          & 
	0.16          &\textit{0.13} & 0.25         \\
	&SVO~\cite{forster2016manifold} & 0.12          & 
	0.16          & X          & 
	X          &X & 0.14+X         \\
	&VI-ORB~\cite{mur2017visual}  & \textit{0.09} & 
	\textbf{0.08} & X           & 
	\textbf{0.04} &\textbf{0.07} & 0.07+X      \\ 
	&VI-DSO~\cite{von2018direct}  & 0.12          & 
	0.12          & \textbf{0.10} & 
	0.06          &\textit{0.17} & \textit{0.11} \\
	\hline
	&\textit{End-end VO}&1.80&0.88&1.00&1.24&0.78&1.14\\
	&\textit{Dd}     & 0.12 & 0.11 & 
	0.63 
	& 0.07 & 0.52 & 0.29 \\
	&\textit{Dd+Dp}    & \textit{0.09} & \textit{0.09} & 
	0.13 
	& 0.06 & 0.19 & 0.11 \\
	&\textit{Dd+Du}     & \textbf{0.08} & \textit{0.09} & 
	0.55 
	& 0.08 & 0.47 & 0.25 \\
	&D3VO    & \textbf{0.08} & \textit{0.09} & \textit{0.11} 
	& \textit{0.05} &0.19 & \textbf{0.10} \\

	\hline
	\hline
	\parbox[t]{2mm}{\ 
		\multirow{3}{1mm}{\rotatebox[origin=c]{90}{\underline{\ \ S+I\ \ 
		}}}}&VINS~\cite{qin2019general}
		  & 0.23          & 
	0.19          & \textit{0.11} & 
	\textit{0.10}          &- & \textit{0.17}         \\
	&OKVIS~\cite{leutenegger2015keyframe} & 0.23          & 
	0.36          & 0.13          & 
	0.17          &- & 0.22          \\
	&Basalt~\cite{usenko2019visual}  & \textcolor{blue}{\textbf{0.06}}          
	& 
	\textit{0.12}          & \textcolor{blue}{\textbf{0.10}}          & 
	\textcolor{blue}{\textbf{0.05}} & - & \textcolor{blue}{\textbf{0.08}} \\ 
	\hline
	&D3VO    & \textit{0.08} & \textcolor{blue}{\textbf{0.09}} & 
	\textit{0.11} 
	& 
	\textcolor{blue}{\textbf{0.05}} &- & \textcolor{blue}{\textbf{0.08}}
\end{tabular}
\caption{Evaluation results on EuRoC 
		MAV~\cite{Burri25012016}. We show the results of DSO and 
		ORB-SLAM as baselines and compare D3VO with other SOTA 
		monocular VIO (M+I) and stereo VIO (S+I) methods. Note that for stereo 
		methods, 
		\textit{V2\_03\_difficult} is excluded due to 
		many missing images from one of the 
		cameras~\cite{usenko2019visual}. Despite being a 
		monocular method, D3VO shows comparable results to 
		SOTA monocular/stereo VIO. The best results among 
		the monocular methods are shown as \textbf{black bold} and the best  
		among the stereo methods are shown as \textcolor{blue}{\textbf{blue 
		bold}}. The ablation study shows that $Dd$+$Dp$ delivers large improvement on 
		\textit{V1\_03\_difficult} and \textit{V2\_03\_difficult} where the 
		camera motions are very strong.}\vspace{-1.5em}
		\label{tab:euroc_vo}
		\vspace{-0.5em}
\end{table}

\begin{figure*}[t]
	\centering
	\includegraphics[width=\linewidth]{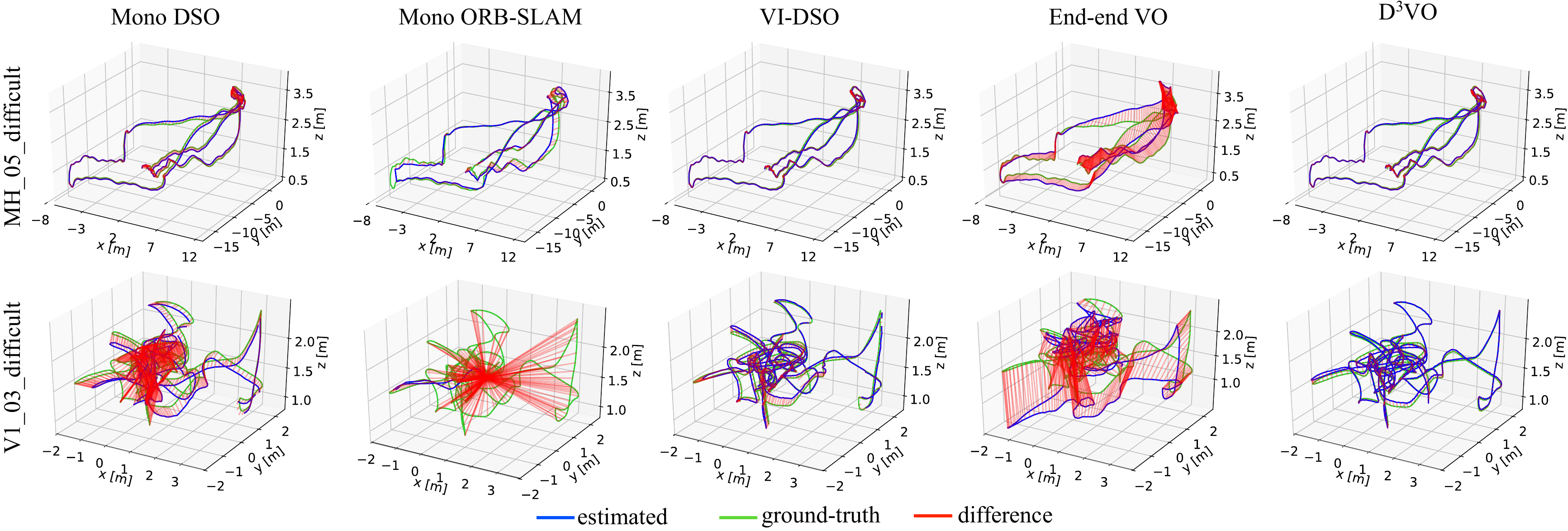}
	\caption{Qualitative comparison of the trajectories on 
	\textit{MH\_05\_difficult} and 
	\textit{V1\_03\_difficult} from EuRoC MAV.}\vspace{-1.2em}
	\label{fig:euroc_traj}
\end{figure*}
\textbf{EuRoC MAV}. As introduced in Sec. 4.1, EuRoC 
MAV is 
very challenging for purely vision-based VO due to the strong 
motion and significant 
illumination changes. VIO 
methods~\cite{von2018direct,usenko2019visual,qin2018vins,leutenegger2015keyframe}
dominate this benchmark by integrating IMU measurements to get a pose or motion prior
and meanwhile estimating the absolute scale. 
We compare D3VO with other 
state-of-the-art monocular VIO (M+I) as well as stereo VIO (S+I) methods on 
sequences
\textit{MH\_03\_medium},  \textit{MH\_05\_difficult}, 
\textit{V1\_03\_difficult}, 
\textit{V2\_02\_medium} and \textit{V2\_03\_difficult}. All the other sequences 
are used for training. We refer to~\cite{delmerico2018benchmark} for 
the results of the M+I methods. The results of DSO and 
ORB-SLAM are shown as 
baselines. We also show the results from the proposed PoseNet \textit{(End-end 
VO}). For the evaluation metric, we use the root mean square (RMS) of the
absolute trajectory error (ATE) after aligning the estimates
with ground truth. The results in Table~\ref{tab:euroc_vo} show 
that with the proposed framework integrating depth, pose and 
uncertainty from the proposed deep neural 
network, D3VO shows high accuracy as well as robustness and is able to 
deliver comparable results to other state-of-the-art VIO methods with only a 
single camera. We also show the ablation study for the integration of predicted 
depth (\textit{Dd}), pose (\textit{Dp}) and uncertainty (\textit{Du}) and the 
integration of pose prediction improves the performance significantly on 
\textit{V1\_03\_difficult} and \textit{V2\_03\_difficult} where violent camera 
motion occurs.

Figure~\ref{fig:euroc_traj} shows the qualitative comparison of trajectories 
obtained from DSO~\cite{engel2017direct}, ORB-SLAM~\cite{mur2015orb}, visual 
inertial 
DSO~\cite{von2018direct}, the 
end-to-end predicted poses from our network 
and D3VO on the \textit{MH\_03} and \textit{V1\_03} 
sequences. All the 5 methods can deliver fairly good results on 
\textit{MH\_05\_difficult}. 
On \textit{V1\_03\_difficult} where the motions are 
stronger and there are many brightness inconsistencies between temporal and 
stereo images, D3VO can still deliver comparable results to 
VI-DSO, while using only a single camera.

\section{Conclusion}
We presented D3VO as a monocular VO method that enhances the
performance of geometric VO methods by exploiting the predictive power
of deep networks on three levels integrating predictions of monocular
depth, photometric uncertainty and relative camera pose.  To this end,
we first introduced a novel self-supervised monocular depth estimation
network which explicitly addresses the illumination change in the
training set with predictive brightness transformation parameters.
The network achieves state-of-the-art results on KITTI and EuRoC MAV.
The predicted depth, uncertainty and pose are then incorporated into
both the front-end tracking and back-end non-linear optimization of a
direct VO pipeline. We systematically evaluated the VO performance of
D3VO on the two datasets. D3VO sets a new state-of-the-art on
KITTI Odometry and also achieves state-of-the-art performance on the
challenging EuRoC MAV, rivaling with leading mono-inertial and
stereo-inertial methods while using only a single camera.

\vspace{3mm}
\noindent
\textbf{Acknowledgements} We would like to thank Niclas Zeller, Lukas 
K{\"o}stler, Oleg Muratov and other 
colleagues from Artisense for their continuous feedbacks. Besides, we would 
like to thank Jakob Engel and Tao Wu for the fruitful discussions during 
the early stages of the project. Last but not least, we also would like to 
thank the reviewers and Klaus H. Strobl for their constructive comments.

\clearpage
{\small
\bibliographystyle{ieee_fullname}
\bibliography{egbib}
}
\clearpage

\section*{Supplementary}
\renewcommand{\thesubsection}{\Alph{subsection}}

\subsection{Network Training Details}
Both DepthNet and PoseNet are implemented with 
PyTorch~\cite{paszke2017automatic} 
and trained on a single Titan X Pascal GPU. We resize the images to $512 \times 
256$ for both KITTI~\cite{Geiger2012CVPR} and EuRoC MAV~\cite{Burri25012016}. 
We use ResNet-18~\cite{he2016deep} as the encoder of DepthNet and it is 
initialized with ImageNet~\cite{russakovsky2015imagenet} pre-trained weights. 
Note that since EuRoC MAV provides grayscale images only, we duplicate 
the images to form 3-channel inputs. 
The decoder of DepthNet and the entire PoseNet are initialized randomly. We use
a batch size of $8$ and the Adam optimizer~\cite{kingma2014adam} with the
number of epochs $20$ and $40$ for KITTI and EuRoC MAV, respectively. The 
learning rate is set to $10^{-4}$ initially and decreased to $10^{-5}$ for the 
last $5$ epochs.

The predicted 
brightness transformation parameters are the same for the 3 channels of the 
input images.
We mask out the over-exposure pixels when applying affine brightness 
transformation, since we found they negatively affect the estimation of the 
brightness parameters. Engel et al. also find similar issues 
in~\cite{engel2015large}.

For the total loss function
\begin{equation}
	L_{total} = \frac{1}{s}\sum_{s} (L_{self}^s + \lambda^{s} 
	L_{reg}^s),
\end{equation}
we use $s = 4$ output scales with and $\lambda^s = 10^{-3} \times 
\frac{1}{2^{s-1}}$. For the regularization
\begin{equation}
L_{reg} = L_{smooth} + \beta L_{ab}
\end{equation}
with
\begin{equation}
	L_{smooth} =  \sum_{\mathbf{p}\in V}\left| 
	\nabla_xD_t\right|e^{-\left|\nabla_xI_t\right|} + \left| 
	\nabla_yD_t\right|e^{-\left|\nabla_yI_t\right|}
\end{equation}
and
\begin{equation}
L_{ab} = \sum_{t'} (a_{t'} - 1)^2 + 
b_{t'}^2,
\end{equation}
we set $\beta = 10^{-2}$.

\subsection{Network Architectures}

\textbf{DepthNet}. We adopt ResNet-18~\cite{he2016deep} as the encoder of 
DepthNet
with the 
implementation from the \textit{torchvision} package in 
PyTorch~\cite{paszke2017automatic}. The decoder architecture is built upon the 
implementation in~\cite{Godard_2019_ICCV} with skip connections from the 
encoder, while the difference is that our 
final outputs contain 3 channels including $D_t$, $D_t^s$ and $\Sigma_t$. 
Table~\ref{tab:depth_arch} shows the detailed architecture of DepthNet decoder.

\textbf{PoseNet}. The architecture of PoseNet is similar 
to~\cite{zhou2017unsupervised} without the explainability mask decoder. PoseNet 
takes 2 channel-wise concatenated images as the input and outputs the relative 
pose and the relative brightness parameters $a$ and $b$. The predicted pose is 
parameterized with translation vector and Euler angles.

\begin{table}[]
	\small
	\begin{tabular}{|l|l|l|l|l|}
		\hline
		\multicolumn{5}{|l|}{\textbf{DepthNet 
		Decoder}}                                         \\ \hline
		\textbf{layer} & \textbf{chns} & \textbf{scale} & \textbf{input}  & 
		\textbf{activation} \\ \hline
		upconv5        & 256           & 32             & econv5          & 
		ELU~\cite{clevert2015fast}                 \\
		iconv5         & 256           & 16             & $\uparrow$upconv5, 
		econv4 & 
		ELU                 \\ \hline
		upconv4        & 128           & 16             & iconv5          & 
		ELU                 \\
		iconv4         & 128           & 8              & $\uparrow$upconv4, 
		econv3 & 
		ELU                 \\
		disp\_uncer4          & \textbf{3}    & 1              & 
		iconv4          & 
		Sigmoid             \\ \hline
		upconv3        & 64            & 8              & iconv4          & 
		ELU                 \\
		iconv3         & 64            & 4              & $\uparrow$upconv3, 
		econv2 & 
		ELU                 \\
		disp\_uncer3          & \textbf{3}    & 1              & 
		iconv3          & 
		Sigmoid             \\ \hline
		upconv2        & 32            & 4              & iconv3          & 
		ELU                 \\
		iconv2         & 32            & 2              & $\uparrow$upconv2, 
		econv1 & 
		ELU                 \\
		disp\_uncer2          & \textbf{3}             & 1              & 
		iconv2          & 
		Sigmoid             \\ \hline
		upconv1        & 16            & 3              & iconv2          & 
		ELU                 \\
		iconv1         & 16            & 1              & 
		$\uparrow$upconv1         & 
		ELU                 \\
		disp\_uncer1          & \textbf{3}    & 1              & 
		iconv1          & 
		Sigmoid             \\ \hline
	\end{tabular}
	\caption{Network architecture of DepthNet decoder. All layers are 
	convolutional 
		layers with kernel size $3$ and stride $1$, and $\uparrow$ is $2 \times 
		2$ 
		nearest-neighbor upsampling. Here \textbf{chns} is the number of 
		output channels, \textbf{scale} is the downscaling factor relative to 
		the input image. Note that the disp\_uncer layers have 3-channel 
		outputs that contain $D_t$, $D_t^s$ and $\Sigma_t$.}
	\label{tab:depth_arch}
\end{table}

\begin{table}[]
	\small
	\begin{tabular}{|l|l|l|l|l|l|l|}
		\hline
		\multicolumn{7}{|l|}{\textbf{PoseNet}}                             \\ 
		\hline
		\textbf{layer}     & \textbf{k} & \textbf{s} & \textbf{chns} & 
		\textbf{scale} & \textbf{input}     & \textbf{activation} \\ \hline
		conv1     & 3 & 2 & 16   & 2     & $I_{t\pm1}$,$I_{t}$       & 
		ReLU       
		\\
		conv2     & 3 & 2 & 32   & 4     & conv1     & 
		ReLU       \\
		conv3     & 3 & 2 & 64   & 8     & conv2     & ReLU       \\
		conv4     & 3 & 2 & 128  & 16    & conv3     & ReLU       \\
		conv5     & 3 & 2 & 256  & 32    & conv4     & ReLU       \\
		conv6     & 3 & 2 & 512  & 64    & conv5     & ReLU       \\
		conv7     & 3 & 2 & 1024 & 128   & conv6     & ReLU       \\
		avg\_pool & - & - & 1024 & -     & conv7     & -          \\ \hline
		pose      & 1 & 1 & 6    & -     & avg\_pool & -          \\ \hline
		a         & 1 & 1 & 1    & -     & avg\_pool & 
		Softplus   \\ \hline
		b         & 1 & 1 & 1    & -     & avg\_pool & TanH       \\ \hline
	\end{tabular}
	\caption{Network architecture of PoseNet. Except for the global average 
	pooling 
	layer (avg\_pool), all layers 
	are convolutional layers with \textbf{k} the kernel size, \textbf{s}
	the stride, \textbf{chns}
	the channels and \textbf{scale} the downscaling 
	factor relative to the input image.}
	\label{tab:pose_arch}
\end{table}

\subsection{Factor Graph of Front-end Tracking}
\begin{figure}
	\includegraphics[width=\linewidth]{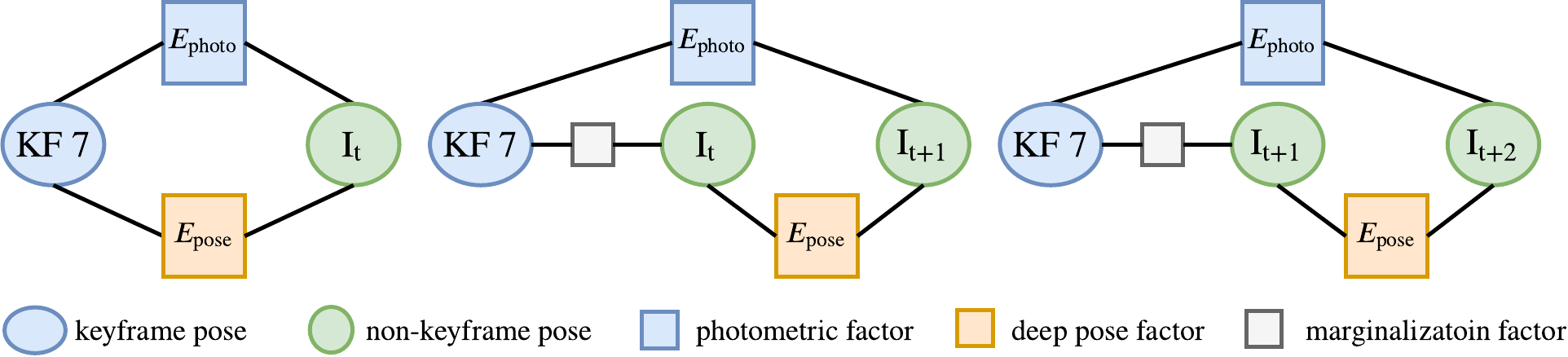}
	\caption{Visualization of the factor graph created for the front-end 
	tracking in D3VO. From left to right are the factor graph when the first 
	($I_t$), second ($I_{t+1}$) and third ($I_{t+2}$) frame comes after the 
	newest keyframe, which is the reference frame for the front-end tracking, 
	is 
	added to the optimization window. The predicted relative poses from the 
	proposed PoseNet is used as the prior between the consecutive frames.}
	\label{fig:fac_g}
\end{figure}

In Figure~\ref{fig:fac_g}, we show the visualization of the factor graphs 
created for the front-end tracking in D3VO. The non-keyframes are tracked with 
respect to the reference frame, which is the latest keyframe in the 
optimization window with direct image alignment. With the predicted relative 
poses from PoseNet, we also add a prior factor between the consecutive frames. 
When the new non-keyframe comes, the oldest non-keyframe in the factor graph is 
marginalized. The figure shows the status of the factor graph for the first 
($I_t$), 
second ($I_{t+1}$) and third non-keyframe ($I_{t+2}$) comes.

\subsection{Additional Experiments on Brightness Parameters}

\begin{table}[t]
	\centering
	\begin{tabular}{|l|c|}
		\hline
		& \multicolumn{1}{l|}{avg photometric error} \\ \hline
		w/o \textit{ab} & 0.10                                       \\ \hline
		w/ \textit{ab}  & \textbf{0.03} \\ \hline
		w/ \textit{ab (LS)}  & \textit{0.07} \\ \hline
	\end{tabular}
	\caption{Average photometric errors on \textit{V2\_03\_difficult}. We 
		project the visible 3D points with ground-truth depth of the left 
		images 
		onto the corresponding right images fo the stereo pairs, and then 
		calculate 
		the absolute photometric errors. Note that the intensity values are 
		normalized 
		to 
		$[0,1]$. 
		The results show that by transforming the left images with the 
		predicted \textit{ab}, the average photometric error is largely 
		decreased.}
	\label{tab:ab_eval}
\end{table}

\begin{figure}
	\centering
	\includegraphics[width=\linewidth]{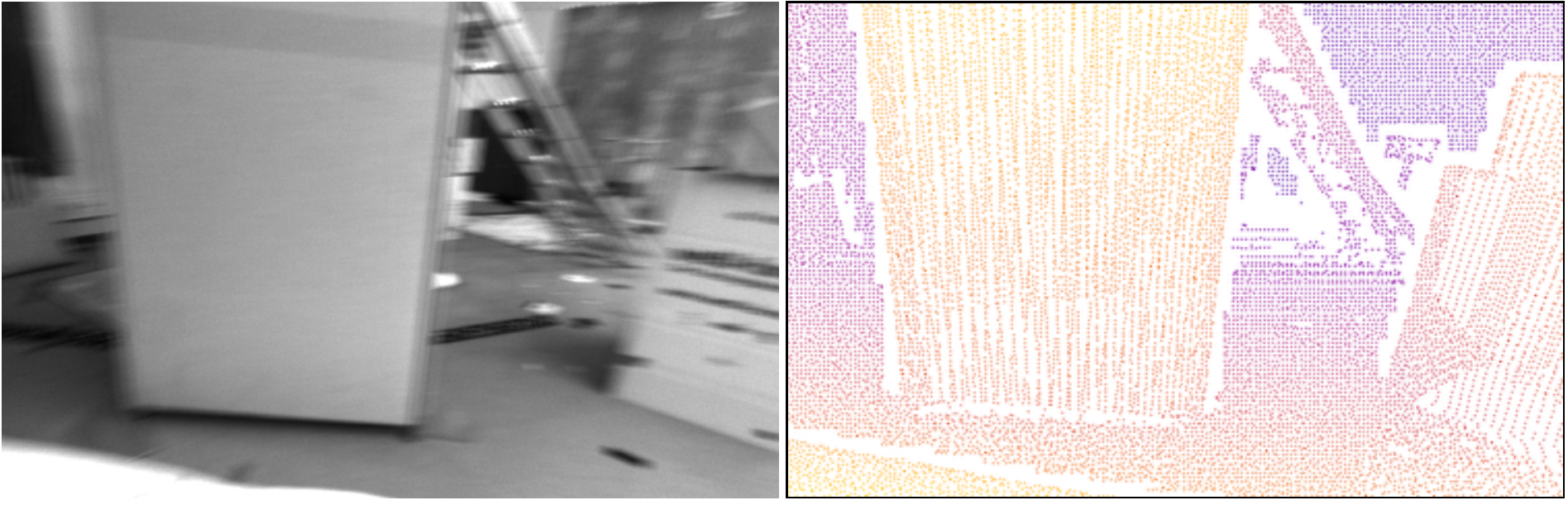}
	\caption{An example of the ground-truth depth map of 
		\textit{V2\_03\_difficult} in EuRoC MAV. 
	}
	\label{fig:gt_depth_euroc}
\end{figure}

\begin{figure}
	\centering
	\includegraphics[width=\linewidth]{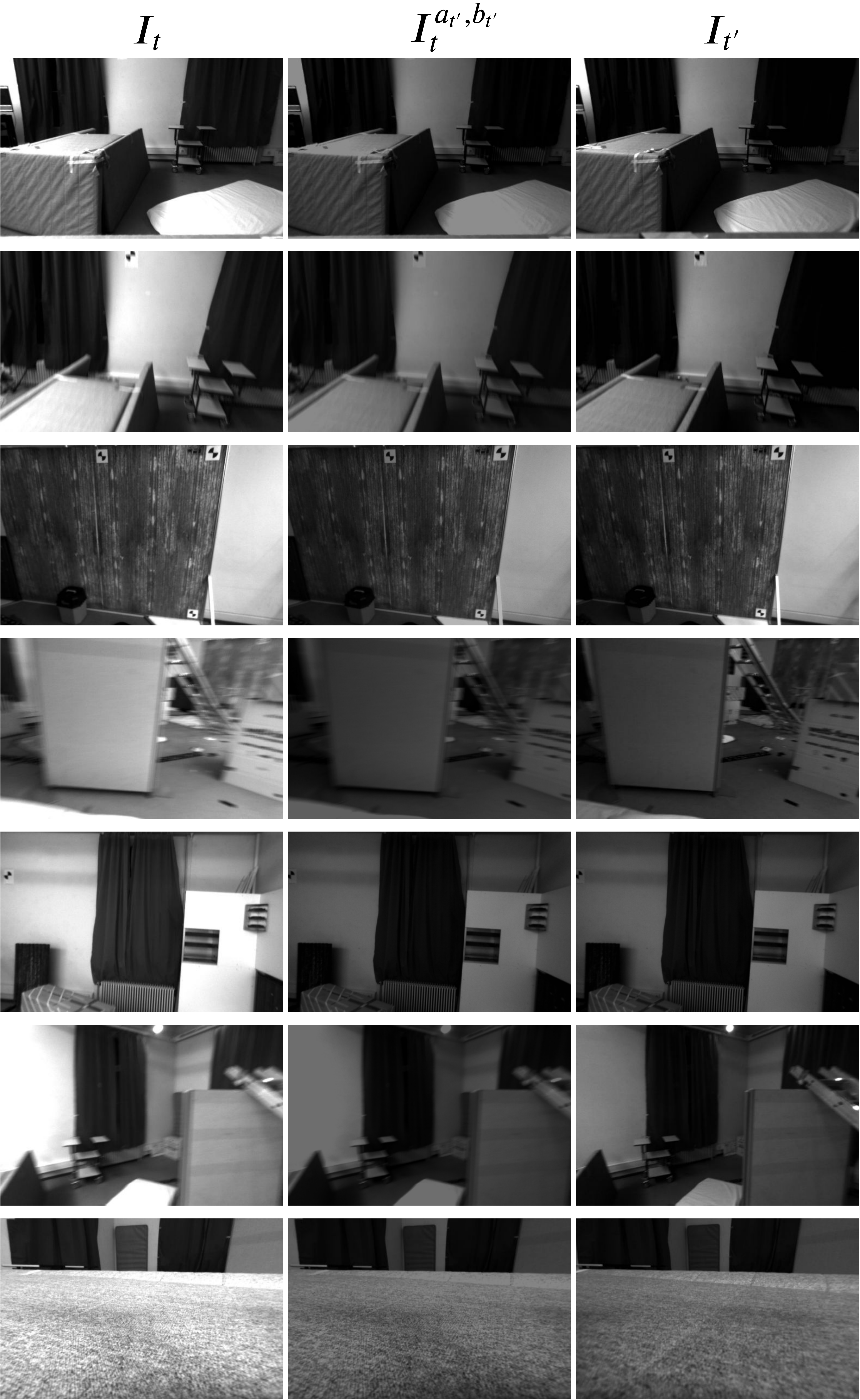}
	\caption{Examples of affine brightness transformation in 
		\textit{V2\_03\_difficult} from EuRoC MAV.}
	\label{fig:supp_aff_euroc}
	\vspace{-1em}
\end{figure}

In our main paper, we have shown that the predictive brightness parameters 
effectively improve the depth estimation accuracy, especially on EuRoC MAV 
where the illumination change is quite strong. To further validate the 
correctness of the predicted brightness parameters, we measure the photometric 
errors when projecting the pixels from the source images to the next 
consecutive images using the ground-truth depth and poses in 
\textit{V2\_03\_difficult}. An example of the ground-truth depth 
is shown in Figure~\ref{fig:gt_depth_euroc} for which we use the code from the 
authors of~\cite{Gordon_2019_ICCV}. We first calculate 
the photometric errors using the original image 
pairs and then calculate the absolute photometric errors by transforming the 
left images 
with the predicted parameters from PoseNet. We also implemented a simple 
baseline method to estimate the affine brightness parameters by solving linear 
least squares (LS). We formulated the normal equation with the dense optical 
flow method~\cite{farneback2003two} implemented in 
OpenCV~\cite{opencv_library}. As shown in 
Table~\ref{tab:ab_eval}, the average photometric error is decreased by a large 
margin when the affine brightness transformation is performed and the predicted 
parameters from PoseNet are better than the ones estimated from LS. We 
show more examples of the affine brightness transformation in 
Figure~\ref{fig:supp_aff_euroc}. 

\newcolumntype{L}[1]{>{\let\newline\\
		\arraybackslash\hspace{0pt}}m{#1}}
\newcolumntype{C}[1]{>{\centering\let\newline\\
		\arraybackslash\hspace{0pt}}m{#1}}
\begin{table}[]
	\small
	\setlength{\tabcolsep}{0.4em}
	\begin{tabular}{L{1.54cm}|C{0.6cm}C{0.6cm}C{0.6cm}C{0.6cm}C{0.6cm}C{0.6cm}|c}
		& 01            & 02            & 06            & 
		08            & 09            & 10            & 
		mean          \\ \hline
		ORB2~\cite{mur2017orb}      & \textbf{21.4}          & 
		\textit{15.0}          & 
		3.52          & \textit{11.1}          & \textit{6.34}          & 
		5.25 & \textit{10.4}          \\
		S. DSO~\cite{wang2017stereoDSO}       & \textit{26.5}          
		& 16.4 & 
		\textit{3.11} & \textbf{11.0} & 9.39          & 
		\textit{3.11} & 11.6         \\ 
		D3VO
		& 26.9         & 
		\textbf{10.4}          & 
		\textbf{2.92}          & 12.7          & \textbf{5.30}          & 
		\textbf{2.44}          & \textbf{10.1}          \\ \hline \hline
		
		ORB2~\cite{mur2017orb}      & 9.95          & 
		9.55         & 
		2.45          & 3.75          & \textit{3.07}          & 
		0.99 & 4.96          \\
		S. DSO~\cite{wang2017stereoDSO}       & \textit{5.08}          
		& \textit{7.82} & 
		\textit{1.93} & \textbf{3.02} & 4.31          & 
		\textit{0.84} & \textit{3.83}         \\ 
		D3VO
		& \textbf{1.73}         & 
		\textbf{5.43}          & 
		\textbf{1.69}          & \textit{3.53}          & 
		\textbf{2.68}          & 
		\textbf{0.87}          & \textbf{2.65}          \\ \hline
	\end{tabular}
	\vspace{-1em}
	\caption{Absolute translational error (ATE) as RMSE on KITTI. The upper 
		part and the lower part show the results w/o and w/ SE(3) alignment, 
		respectively. Note that ATE 
		is very sensitive to the error occurs at one specific 
		time~\cite{zhang2018tutorial}.}
	\label{tab:kitti_ate}
	\vspace{-1em}
\end{table}

\begin{figure}
	\centering
	\includegraphics[width=.45\linewidth]{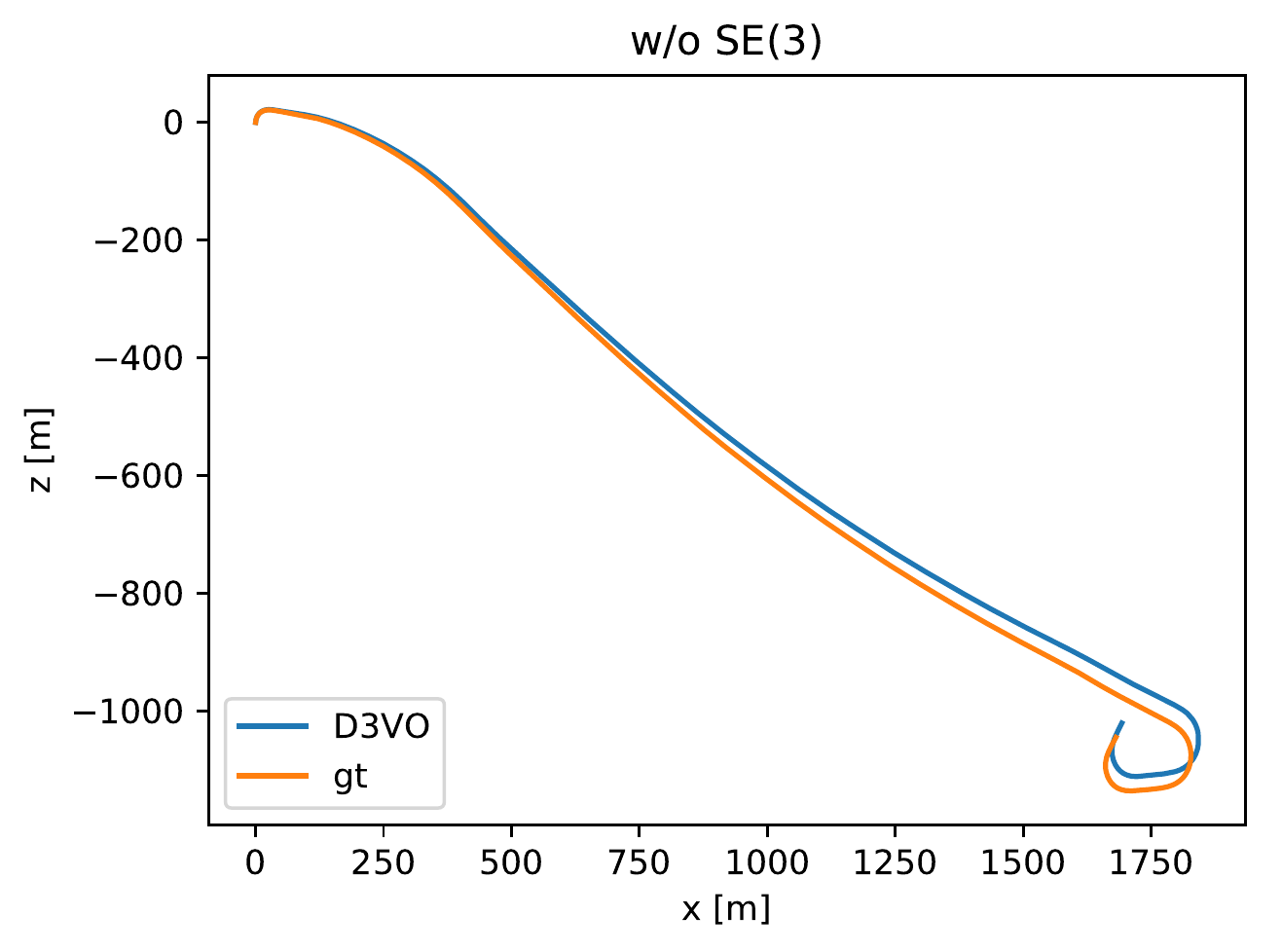}
	\includegraphics[width=.45\linewidth]{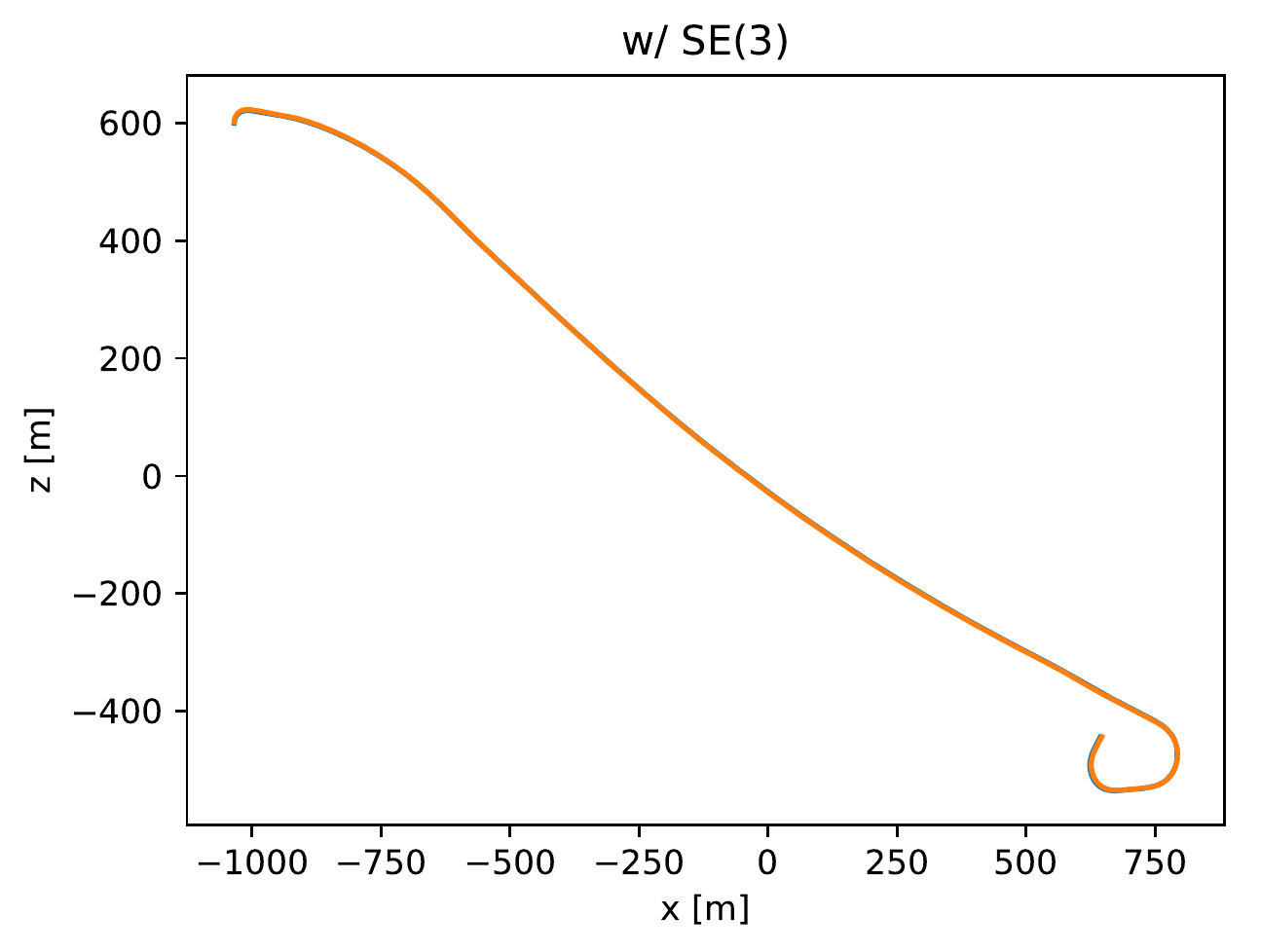}
	\includegraphics[width=.45\linewidth]{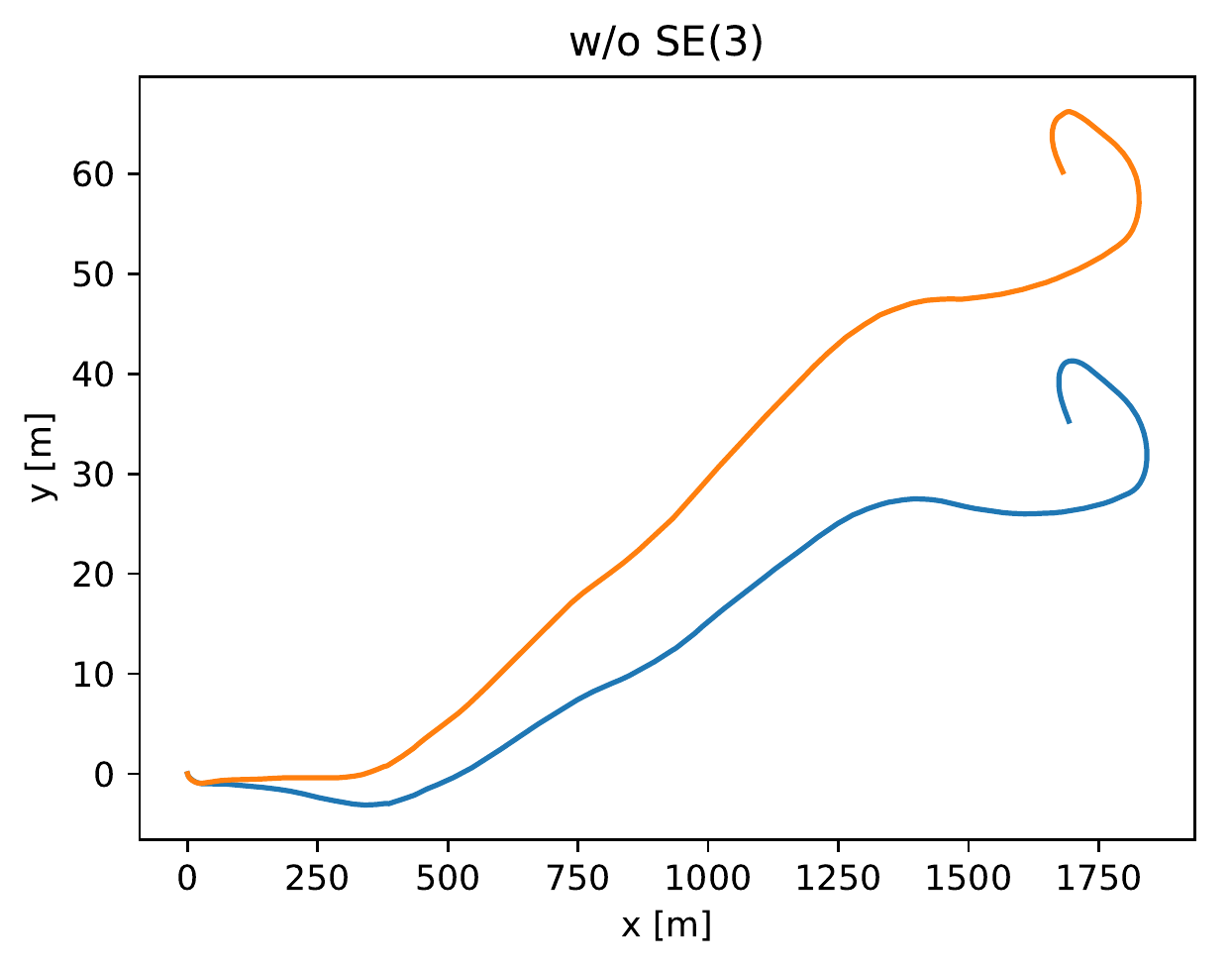}
	\includegraphics[width=.45\linewidth]{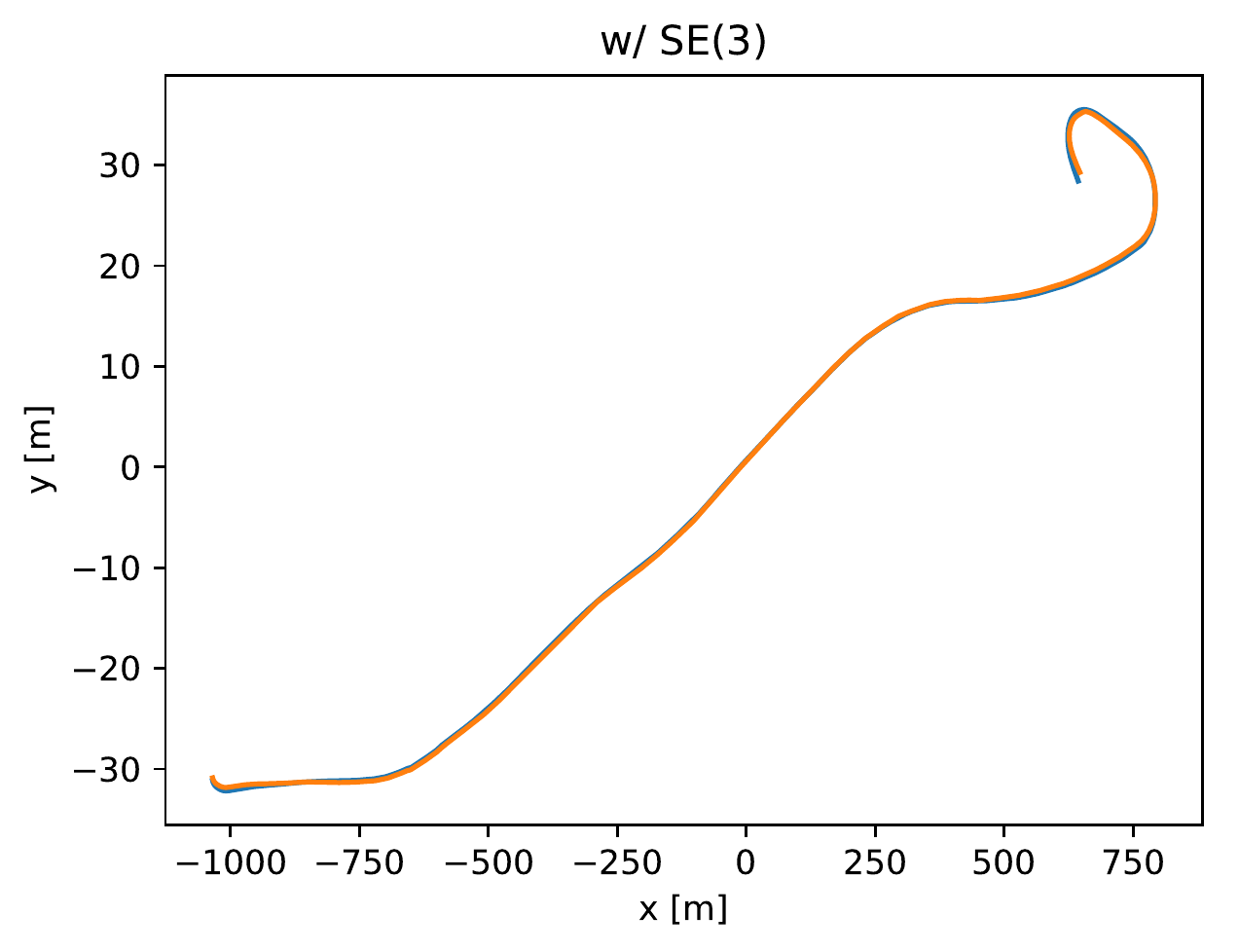}
	\caption{Trajectories on KITTI 01 to compare between w/o and w/ SE(3) 
	alignment for the ATE evaluation. The upper part of the figure shows the 
	trajectories on the x-z 
	plane and the lower part shows the trajectories on the x-y plane.  We can 
	see that less accurate pose estimations for the initial frames may result 
	in a large overall ATE, if no SE(3) alignment is performed.}
	\label{fig:ate}
	\vspace{-1em}
\end{figure}

\subsection{Absolute Translational Error on KITTI}
The evaluation metrics proposed with the KITTI benchmark~\cite{Geiger2012CVPR} 
measures the relative pose accuracy. It is important to measure the global 
consistency of the pose estimations. Therefore, we also show the absolute 
translational error (ATE) as RMSE in Table~\ref{tab:kitti_ate} where the upper 
part shows the evaluation results without the SE(3) alignment and the lower 
part shows the results with the SE(3) alignment. For some sequences, e.g., 
KITTI~01, the ATE without SE(3) alignment is very large, while the ATE with 
SE(3) alignment dramatically decreases. The trajectories on KITTI~01 are shown 
in Figure.~\ref{fig:ate} where we can see that the less accurate pose 
estimations for the initial frames may result in a large overall ATE.

\subsection{Cityscapes}

\begin{figure}
	\centering
	\includegraphics[width=\linewidth]{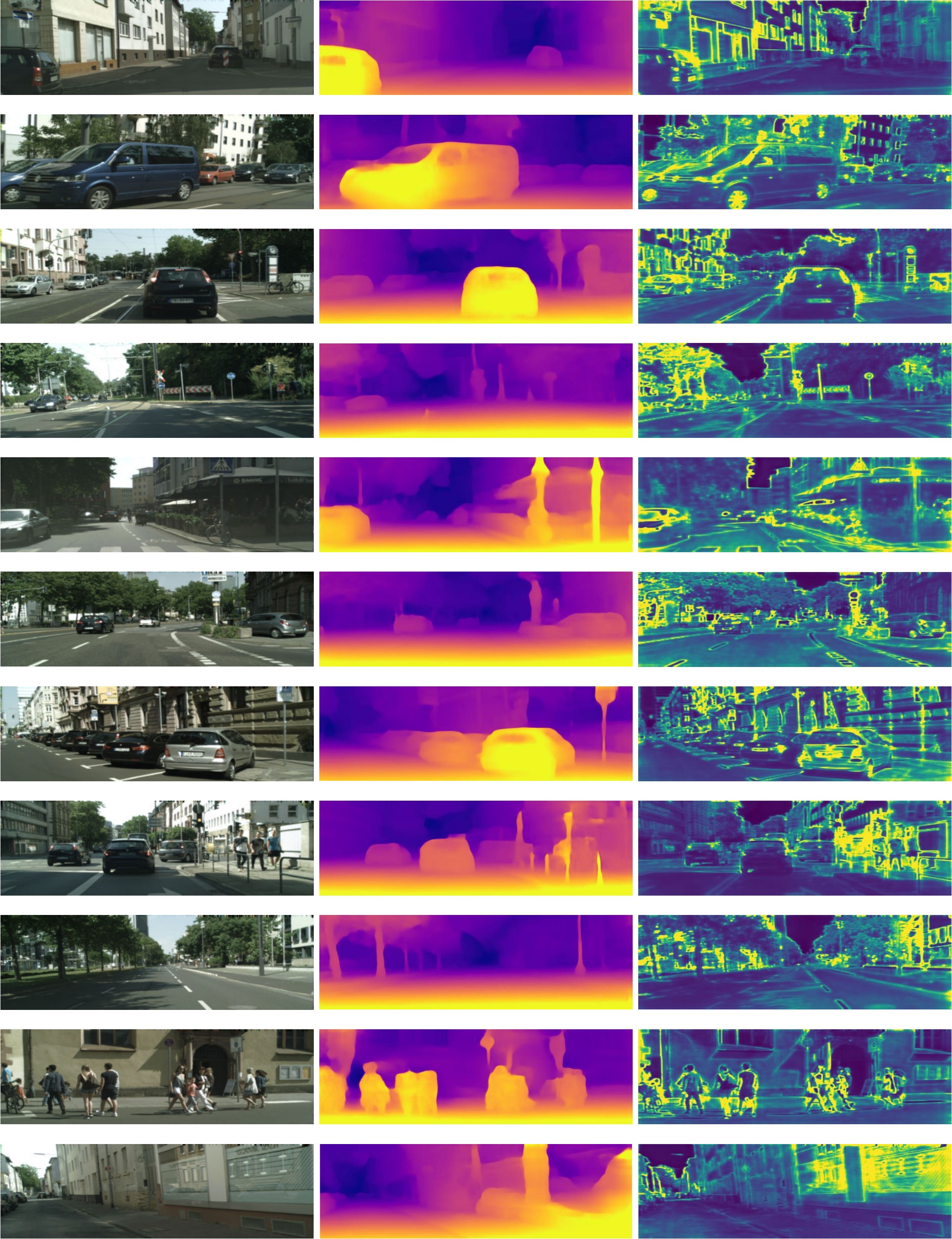}
	\caption{Results on Cityscapes with the model trained on KITTI.}
	\label{fig:supp_cs}
\end{figure}

Figure~\ref{fig:supp_cs} shows the results on the Cityscapes 
dataset~\cite{cordts2016cityscapes} with our 
model trained on KITTI. The results show the generalization capability of our 
network on both depth and uncertainty prediction. In 
particular, the network can generalize to predict high uncertainties on 
reflectance, object boundaries, high-frequency areas, and moving objects.

\end{document}